\documentclass[10pt,journal,compsoc]{IEEEtran}
\usepackage{ifpdf}
\ifCLASSOPTIONcompsoc
  \usepackage[nocompress]{cite}
\else
  \usepackage{cite}
\fi
\ifCLASSINFOpdf
  \usepackage[pdftex]{graphicx}
\else
  \usepackage[dvips]{graphicx}
\fi
\usepackage{amsmath}
\usepackage{algorithmic}
\usepackage{array}
\ifCLASSOPTIONcompsoc
 \usepackage[caption=false,font=footnotesize,labelfont=sf,textfont=sf]{subfig}
\else
 \usepackage[caption=false,font=footnotesize]{subfig}
\fi
\usepackage{url}
\usepackage{color}
\usepackage{amsfonts}
\usepackage{multirow}
\usepackage[ruled,vlined]{algorithm2e}
\usepackage{tikz}

\newcommand{\ie}{\emph{i.e.}}
\newcommand{\eg}{\emph{e.g.}}
\newcommand{\etal}{\emph{et.\,al.}}
\newcommand{\etc}{\emph{etc}}

\begin{document}
\title{Visibility Constrained Generative Model for Depth-based 3D Facial Pose Tracking}

\author{Lu~Sheng,~\IEEEmembership{Member,~IEEE}
        Jianfei~Cai,~\IEEEmembership{Senior~Member,~IEEE}
        Tat-Jen~Cham,~\IEEEmembership{}
        Vladimir~Pavlovic,~\IEEEmembership{Senior~Member,~IEEE}
        and~King~Ngi~Ngan,~\IEEEmembership{Fellow,~IEEE}% <-this % stops a space
\IEEEcompsocitemizethanks{
\IEEEcompsocthanksitem L. Sheng is with the College of Software, Beihang University, China.\protect\\
E-mail: lsheng@buaa.edu.cn
\IEEEcompsocthanksitem K. N. Ngan is with University of Electronic Science and Technology.\protect\\
E-mail: knngan@ee.cuhk.edu.hk
\IEEEcompsocthanksitem J. Cai and T-J. Cham are with the School of Computer Science and Engineering, Nanyang Technological University, Singapore.\protect\\
E-mail: \{asjfcai, astjcham\}@ntu.edu.sg
\IEEEcompsocthanksitem V. Pavlovic is with the Department of Computer Science, Rutgers University, USA.\protect\\
E-mail: vladimir@cs.rutgers.edu}% <-this % stops an unwanted space
}


\markboth{Preprint}%
{Sheng \MakeLowercase{\textit{et al.}}:Visibility Constrained Generative Model for Depth-based 3D Facial Pose Tracking}

% \IEEEpubid{0000--0000/00\$00.00~\copyright~2015 IEEE}

% \IEEEspecialpapernotice{(Invited Paper)}

\IEEEtitleabstractindextext{%
\begin{abstract}

In this paper, we propose a generative framework that unifies depth-based 3D facial pose tracking and face model adaptation on-the-fly, in the unconstrained scenarios with heavy occlusions and arbitrary facial expression variations. 
Specifically, we introduce a statistical 3D morphable model that flexibly describes the distribution of points on the surface of the face model, with an efficient switchable online adaptation that gradually captures the identity of the tracked subject and rapidly constructs a suitable face model when the subject changes.
Moreover, unlike prior art that employed ICP-based facial pose estimation, to improve robustness to occlusions, we propose a ray visibility constraint that regularizes the pose based on the face model's visibility with respect to the input point cloud.
Ablation studies and experimental results on Biwi and ICT-3DHP datasets demonstrate that the proposed framework is effective and outperforms completing state-of-the-art depth-based methods.

\end{abstract}

\begin{IEEEkeywords}
3D facial pose tracking, generative model, depth, online Bayesian model, mixture of Gaussian models
\end{IEEEkeywords}}

\maketitle

\IEEEdisplaynontitleabstractindextext

% \fi
\IEEEpeerreviewmaketitle

\IEEEraisesectionheading{\section{Introduction}\label{sec:introduction}}

\IEEEPARstart{R}{obust} 3D facial pose tracking is a central task in many computer vision and computer graphics problems, with applications in facial performance capture, human-computer interaction, as well as VR/AR applications in modern mobile devices.
Although the facial pose tracking has been successfully performed on RGB data~\cite{cao20133d,Egger2018,Booth_2017_CVPR,Jackson_2017_ICCV,Richardson_2017_CVPR,saito2016real,Dou_2017_CVPR,cao2014displaced,cao2015real,blanz1999morphable} for well-constrained scenes, challenges posed by illumination variations, shadows, and substantial occlusions make these RGB-based facial pose tracking approaches less reliable in \emph{unconstrained} scenarios.
The utilization of depth data from commodity real-time range sensors has led to more robust 3D facial pose tracking, not only by enabling registration in the 3D metric space but also by providing cues for the occlusion reasoning.

\begin{figure}
\centering
\includegraphics[width=\linewidth]{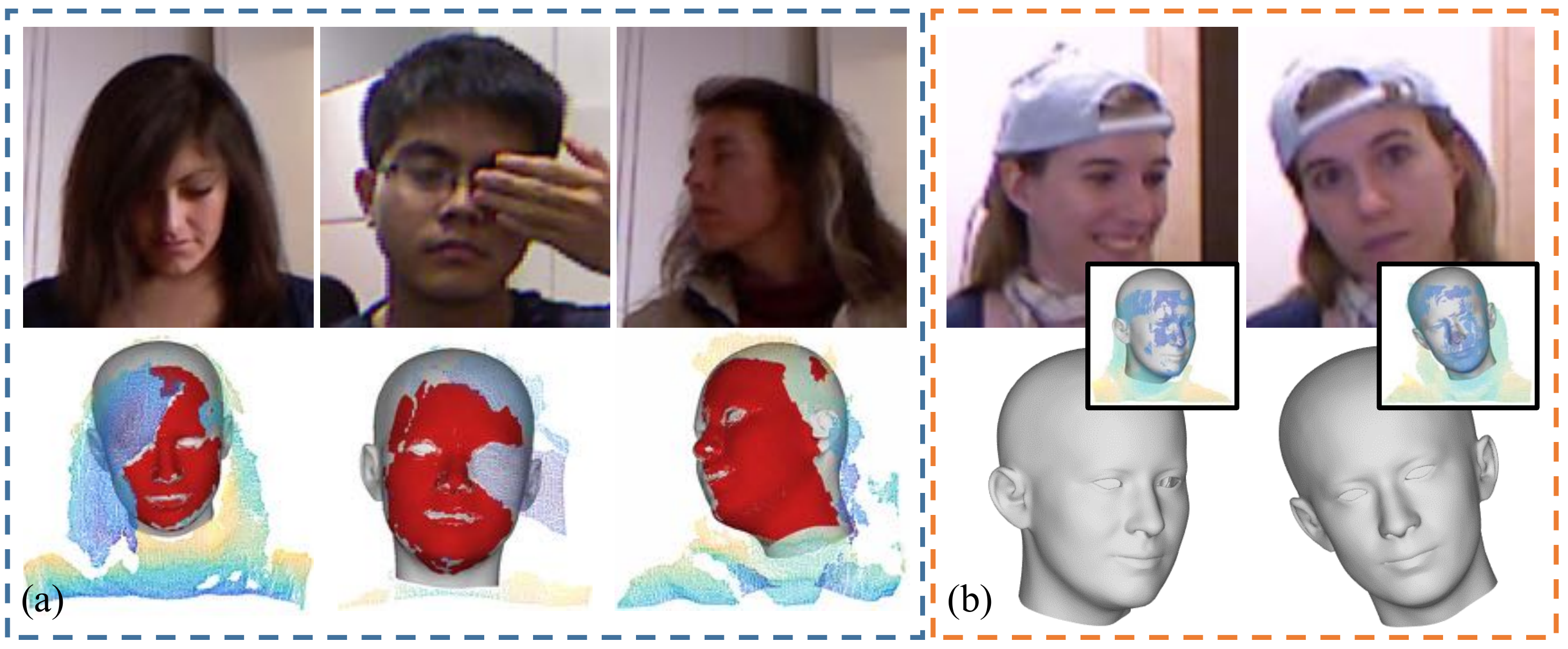}
\caption{Our identity-adaptive facial pose tracking system is robust to occlusions and expressions. (a) Poses are estimated under heavy occlusions. The face models are overlaid with the point clouds. The visible face regions marked in red. (b) Poses are tracked under varying expressions. The estimated face identities are not interfered by expressions.}
\label{fig:introduction}
\end{figure}

Although promising results have been demonstrated by leveraging both RGB and depth data~\cite{hsieh2015unconstrained,pham2016robust,thomas2016augmented,weise2011realtime,li2013realtime,8368264,tan2017facecollage}, or even RGB data alone~\cite{saito2016real,Egger2018} in unconstrained facial pose tracking, existing approaches are not yet able to reliably cope with RGB data affected by inconsistent or poor lighting conditions.
For example, mobile applications like FaceID or Animoji (which heavily employ the face tracking module in their systems) are bound to fail in dark scenes such as bedrooms and theaters, or scenes under complex illuminations such as parties and clubs.
Furthermore, RGB data may be deliberately suppressed in scenarios where privacy is a major concern.
Therefore, it is meaningful to study robust 3D facial pose tracking using depth data alone, complementary to traditional RGB-based tracking systems.

Several key challenges need to be addressed in the context of depth-based tracking:
(1) coping with complex self-occlusions and other occlusions caused by hair, accessories, hands and \etc.;
(2) sustaining an always-on face tracker that can dynamically adapt to any user without manual recalibration;
and (3) providing stability over time to variations in user expressions. 
Unlike previous depth-based discriminative or data-driven methods~\cite{sun2008automatic,breitenstein2008real,papazov2015real,fanelli2011real,fanelli2011randomforests,riegler2014hough,kazemi2014real} that require complex training or manual calibration, in this paper we propose a framework that unifies pose tracking and face model adaptation on-the-fly, offering highly accurate, occlusion-aware and uninterrupted 3D facial pose tracking, as shown in Fig.~\ref{fig:introduction}.
The contributions of this work are fourfold:

1) We introduce a decomposable statistical formulation for a 3D morphable face model, improving upon the earlier 3DMM models~\cite{cao2014facewarehouse,vlasic2005face}.
This formulation encourages a group-wise pose estimation for any potential face model and enables expression-invariant face model updating.

2) We propose an occlusion-aware pose estimation mechanism based on minimizing an information-theoretic \emph{ray visibility score} that optimizes the visibility of our statistical face model.
The mechanism is based on the intuition that valid poses imply that the face points must either be co-located (\ie, \emph{visible}) with the observed point cloud or be \emph{occluded} by the point cloud.
 Without any need for explicit correspondences, our method is highly effective in handling various types of occlusions.

3) We introduce a flow-based constraint to enforce temporal coherence among poses of neighboring frames, in which the poses are regularized by per-pixel depth flows between adjacent frames rather than predicting the motion patterns from the previous pose trajectories.

4) We present an online switchable identity adaptation approach to gradually adapt the face model to the captured subject, or instantaneously switch among stored personalized identity models and create novel face models for new identities.

We present a comprehensive ablation study for the proposed facial tracking framework.
Moreover, experiments on Biwi and ICT-3DHP datasets manifest the superiority of the proposed method against competing depth-based tracking systems.
Note that an early version of this work was published in~\cite{Sheng_2017_CVPR}.
Compared to it, this paper has made substantial extensions including a flow-based temporal coherence, an online switchable identity adaptation method, and a comprehensive ablation study.
With these new technical changes, our proposed framework significantly improves the results of our early method, outperforming the state-of-the-art methods on two benchmark datasets.

\section{Related Work}
\label{sec:related_work}

Facial pose tracking and model regression methods typically consider the RGB videos as their input modality.
They largely rely on tracking the dynamics of sparse 3D facial features that correspond to parametric 3D face models~\cite{Li1993motion,Black1995tracking,DeCarlo2000opticalflow,blanz1999morphable,cao20133d,cao2014displaced}.
In the presence of reliable feature detection, the facial pose can be tracked accurately under moderate occlusions and smooth motion patterns.
Recent advances in discriminative pose estimation and face model reconstruction, which employ deep learning~\cite{Richardson_2017_CVPR,Dou_2017_CVPR,Jackson_2017_ICCV} or random forest~\cite{kazemi2014one,cao20133d,cao2014displaced,guo2018cnn} paradigms, have shown promising results in many applied scenarios.
To improve robustness of prior work, explicit modeling of the occlusions has also been considered~\cite{kazemi2014one,Egger2018,saito2016real}.

Leveraging the introduction and development of depth sensors, a variety of depth-based 3D facial pose tracking and model personalization frameworks have been proposed.
One category of approaches employ sparse depth features, such as facial surface curvatures~\cite{sun2008automatic}, nose tips~\cite{breitenstein2008real}, or triangular surface patch descriptors~\cite{papazov2015real} as the means for a robust 3D facial pose estimation. 
However, these methods may fail when such features cannot be detected under conditions of highly noisy depth data, extreme poses or large occlusions.
A different family of approaches considers discriminative methods based on random forests~\cite{fanelli2011real,fanelli2011randomforests}, deep Hough network~\cite{riegler2014hough}, or finding a dense correspondence field between the input depth image and a predefined canonical face model~\cite{kazemi2014real,wei2016dense}.
Although promising and often accurate, these methods require sophisticated supervised training with large-scale, tediously labeled datasets.

A different modeling strategy involves rigid and non-rigid registration of 3D face models to the depth images, either through the use of 3D morphable models~\cite{cao2014facewarehouse,chen2014depth,pham2016robust,brunton2014review,bouaziz2013online,hsieh2015unconstrained,meyer2015robust,li2015real,storer20093d,6313594,8010438,Egger2018}, or brute-force per-vertex 3D face reconstruction~\cite{tulyakov2014robust,weise2011realtime,li2013realtime,8368264,weise2011realtime,cao2014facewarehouse}.
Although such systems may be accurate, they often require offline initialization or user calibration to create face models specific to individual users.
Subsequent works have been developed to gradually refine the 3D morphable model over time during active tracking~\cite{li2015real,meyer2015robust,bouaziz2013online,hsieh2015unconstrained,thomas2016augmented,li2013realtime}.
Our proposed method falls into this category.
Inspired by statistical models~\cite{cao2014facewarehouse,vlasic2005face,Egger2018,8010438}, we enhance this face model through a decomposable statistical formulation, in which the shape variations from identity and expression are explicitly disentangled.

Occlusion handling is vital for robust 3D facial pose tracking.
While occlusions may be elucidated through face segmentation~\cite{hsieh2015unconstrained,saito2016real,Egger2018} or patch-based feature learning~\cite{fanelli2011randomforests,fanelli2011real,riegler2014hough,kazemi2014real}, iteratively closest points (ICP) based face model registration frameworks do not handle ambiguous correspondences well~\cite{rusinkiewicz2001efficient,hsieh2015unconstrained,weise2011realtime,li2013realtime}. 
Possible remedies include particle swarm optimization~\cite{padeleris2012head} for optimizing complex objective functions~\cite{meyer2015robust}.
Recently, Wang~\etal~\cite{wang2016capturing} tackled for partial registration of general moving subjects and improved occlusion handling by considering multi-view visibility consistency.
Our proposed ray visibility score incorporates a similar visibility constraint between the face model and the input point cloud but with a probabilistic formulation, which is able to more robustly handle uncertainties in the 3D face model, and is thus less vulnerable to local minima that are frequently encountered in ICP.

Our online switchable identity adaptation falls in the category of adaptive online approaches~\cite{hsieh2015unconstrained,thomas2016augmented,li2013realtime,bouaziz2013online,meyer2015robust,Sheng_2017_CVPR}.  However, unlike prior work we also tackle the important problem of identity instantiation and switching, which enables us to rapidly adapt to new users.

\section{Probabilistic 3D Face Parameterization}
\label{sec:probabilistic_3d_face_parameterization}

In this section, we introduce the 3D morphable face model with a probabilistic interpretation, which acts as an effective prior for facial pose estimation and face identity adaptation.

\subsection{Multilinear Face Model}
\label{sub:multilinear_face_model}

3D face shapes are usually represented by triangular meshes, but in this paper we only focus on the parameterization of vertices, while leaving the edges unchanged.
Thus the face shape can be simplified as a vector constructed by an ordered 3D vertex list $\mathbf{f} = [x_1, y_1, z_1, \ldots, x_{N_\mathcal{M}}, y_{N_\mathcal{M}}, z_{N_\mathcal{M}}]^\top$, where the $n^\text{th}$ vertex is $[x_n, y_n, z_n]^\top \in \mathbb{R}^3, \forall n \in \{1, \ldots, N_\mathcal{M}\}$.
$N_\mathcal{M}$ is the total number of vertices in the model.

We apply the multilinear model~\cite{cao2014facewarehouse,vlasic2005face} to parametrically generate $3$D faces that are adaptive to different identities and expressions. 
It is controlled by a three dimensional tensor $\mathcal{C} \in \mathbb{R}^{3N_\mathcal{M} \times N_\text{id} \times N_\text{exp}}$, where the dimensions correspond to shape, identity and expression, respectively.
Thus, the multilinear model represents a 3D face shape as
\begin{equation}
\mathbf{f} = \bar{\mathbf{f}} + \mathcal{C}\times_2 \mathbf{w}_\text{id}^\top \times_3 \mathbf{w}_\text{exp}^\top, \label{eq:face_parameterization}
\end{equation}
where $\mathbf{w}_\text{id}\in \mathbb{R}^{N_\text{id}}$ and $\mathbf{w}_\text{exp}\in \mathbb{R}^{N_\text{exp}}$ are the linear weights for identity and expression, respectively. 
$\times_i$ denotes the $i$-th mode product.
$\bar{\mathbf{f}}$ is the mean face in the training dataset.
The tensor $\mathcal{C}$, also called the core tensor encoding the subspaces of the shape variations in faces, is calculated by high-order singular value decomposition (HOSVD) onto the training dataset, \ie, $\mathcal{C} = \mathcal{T}\times_2 \mathbf{U}_\text{id} \times_3 \mathbf{U}_\text{exp}$. 
$\mathbf{U}_\text{id}$ and $\mathbf{U}_\text{exp}$ are unitary matrices of the mode-$2$ and mode-$3$ HOSVD of the 3D data tensor $\mathcal{T}$.
$\mathcal{T}$ stacks the mean-subtracted face offsets from the training dataset, along the identity and expression dimensions, respectively.

\subsection{Statistical Modeling of the Multilinear Face Model}
\label{sub:proposed_statistical_face_model}

Unlike using a deterministic face template to match the target point cloud, we apply a statistical model where the potential face shape varies inside a learned shape uncertainty around the mean face, thus we have a better chance to find a suitable face prototype compatible with the target point cloud.
Such a model provides a probabilistic prior for robust face pose tracking.

\subsubsection{Identity and Expression Priors} 
\label{sub:identity_and_expression_priors}

According to Eq.~\ref{eq:face_parameterization}, the multilinear model is controlled by the identity weight $\mathbf{w}_\text{id}$ and expression weight $\mathbf{w}_\text{exp}$.
It is convenient to assume that $\mathbf{w}_\text{id}$ and $\mathbf{w}_\text{exp}$ both follow Gaussian distributions: $\mathbf{w}_\text{id} = \boldsymbol\mu_{\text{id}} + \boldsymbol\epsilon_{\text{id}}, \boldsymbol\epsilon_{\text{id}} \sim \mathcal{N}(\boldsymbol\epsilon_{\text{id}} | \mathbf{0}, \boldsymbol\Sigma_\text{id})$ and $\mathbf{w}_\text{exp} = \boldsymbol\mu_{\text{exp}} + \boldsymbol\epsilon_{\text{exp}}, \boldsymbol\epsilon_{\text{exp}} \sim \mathcal{N}(\boldsymbol\epsilon_{\text{exp}} | \mathbf{0}, \boldsymbol\Sigma_\text{exp})$. 

Notice that $\boldsymbol\mu_\text{id}$ (or $\boldsymbol\mu_\text{exp}$) should not be $\mathbf{0}$ as it will possibly make the face model $\mathbf{f}$ insensitive to $\mathbf{w}_\text{exp}$ (or $\mathbf{w}_\text{id}$)~\cite{bolkart2013statistical}.
If we assume either $\boldsymbol\mu_\text{id}$ (or $\boldsymbol\mu_\text{exp}) \simeq \mathbf{0}$, the variation of the expression (or identity) parameters will be less significant for forming the face shape, \ie, $\mathcal{C}\times_2 \mathbf{w}_\text{id}^\top \times_3 \mathbf{w}_\text{exp}^\top \simeq \mathbf{0}$.

\subsubsection{Multilinear Face Model} 
\label{ssub:multilinear_face_model}

\begin{figure}[t]
\centering
\includegraphics[width=\linewidth]{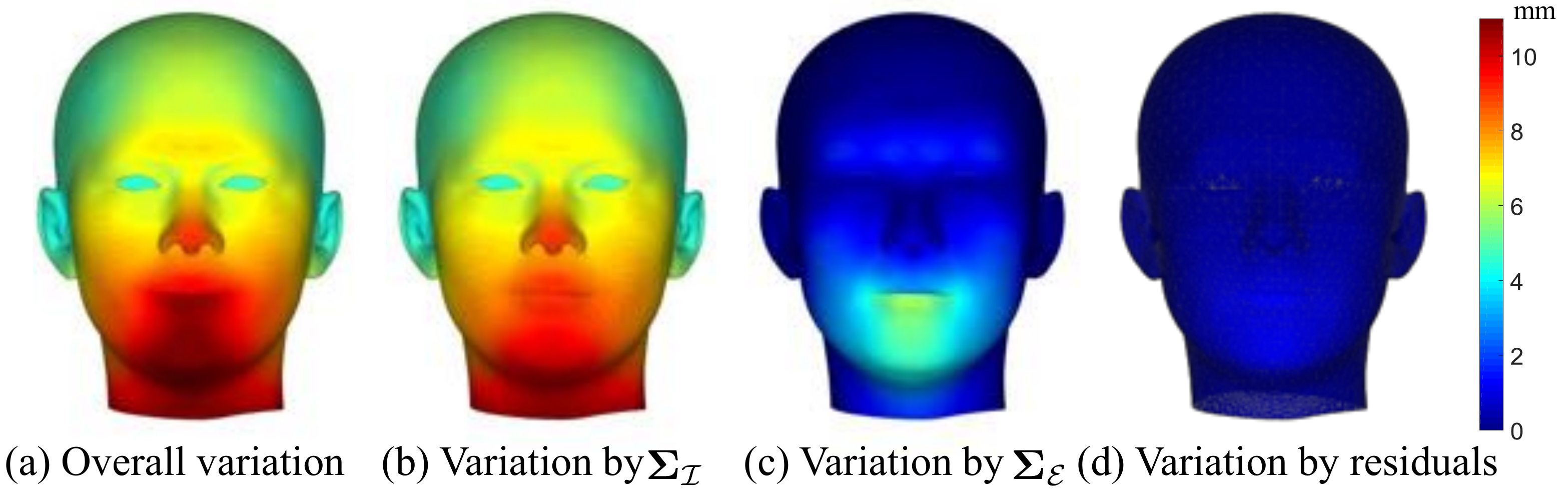}
\caption{The statistics of the face model trained in the FaceWarehouse dataset~\cite{cao2014facewarehouse}. (a) Overall shape variation. (b)--(c) Shape variations by $\mathbf{w}_\text{id}$ and $\mathbf{w}_\text{exp}$, respectively. (d) Shape variation by the residual term in Eq.~\eqref{eq:face_model_with_random_variables}. The shape variation is set as one standard deviation of the marginalized per-vertex distribution.}
\label{fig:multilinear_face_model}
\end{figure}

The canonical face model $\mathcal{M}$ with respect to $\mathbf{w}_\text{id}$ and $\mathbf{w}_\text{exp}$ can be written by re-organizing Eq.~\eqref{eq:face_parameterization}, as
\begin{equation}
\begin{split}
\mathbf{f} = \bar{\mathbf{f}} + \mathcal{C} \times_2 \boldsymbol\mu_\text{id} \times_3 \boldsymbol\mu_\text{exp}  + \mathcal{C} \times_2 \boldsymbol\epsilon_\text{id} \times_3 \boldsymbol\mu_\text{exp} \\
+ ~\mathcal{C} \times_2 \boldsymbol\mu_\text{id} \times_3 \boldsymbol\epsilon_\text{exp} + \mathcal{C} \times_2 \boldsymbol\epsilon_\text{id} \times_3 \boldsymbol\epsilon_\text{exp}. 
\end{split}
\label{eq:face_model_with_random_variables}
\end{equation}
The last term in~\eqref{eq:face_model_with_random_variables} is usually negligible in the shape variation, as illustrated in Fig.~\ref{fig:multilinear_face_model}.
Therefore, $\mathcal{M}$ approximately follows a Gaussian distribution as
\begin{equation}
p_\mathcal{M}(\mathbf{f}) = \mathcal{N}(\mathbf{f}|\boldsymbol\mu_\mathcal{M}, \boldsymbol\Sigma_\mathcal{M}),\label{eq:face_model_marginalization}
\end{equation}
where its mean face shape is 
$\boldsymbol\mu_\mathcal{M} = \bar{\mathbf{f}} + \mathcal{C}\times_2\boldsymbol\mu_\text{id} \times_3 \boldsymbol\mu_\text{exp}$, 
and its variance matrix is
$\boldsymbol\Sigma_\mathcal{M} = \mathbf{P}_\text{id}\boldsymbol\Sigma_\text{id}\mathbf{P}_\text{id}^\top + \mathbf{P}_\text{exp}\boldsymbol\Sigma_\text{exp}\mathbf{P}_\text{exp}^\top$.
The projection matrices $\mathbf{P}_\text{id}$ and $\mathbf{P}_\text{exp}$ for identity and expression are defined as 
$\mathbf{P}_\text{id} = \mathcal{C}\times_3\boldsymbol\mu_\text{exp} \in \mathbb{R}^{3N_\mathcal{M}\times N_\text{id}}$ and $
\mathbf{P}_\text{exp} = \mathcal{C}\times_2\boldsymbol\mu_\text{id} \in \mathbb{R}^{3N_\mathcal{M}\times N_\text{exp}}$, respectively.

Since we are also interested in estimating the identity distribution for personalization of the face shape, we convert the canonical face distribution into a joint distribution for the face shape $\mathbf{f}$ and the identity parameter $\mathbf{w}_\text{id}$, as
\begin{align}
p(\mathbf{f}, \mathbf{w}_\text{id}) & = p_\mathcal{M}(\mathbf{f}|\mathbf{w}_\text{id})p(\mathbf{w}_\text{id}) \notag \\
& = \mathcal{N}(\mathbf{f}|\bar{\mathbf{f}} + \mathbf{P}_\text{id}\mathbf{w}_\text{id}, \boldsymbol\Sigma_\mathcal{E})\mathcal{N}(\mathbf{w}_\text{id}|\boldsymbol\mu_\text{id}, \boldsymbol\Sigma_\text{id}), \label{eq:face_model_identity}
\end{align}
where the expression variance is $\boldsymbol\Sigma_\mathcal{E} = \mathbf{P}_\text{exp}\boldsymbol\Sigma_\text{exp}\mathbf{P}_\text{exp}^\top$.

As shown in Fig.~\ref{fig:multilinear_face_model}, the overall shape variation (represented as per-pixel standard deviation) is, unsurprisingly, most significant in the facial region as compared to other parts of the head.
We further observe that this shape variation is dominated by differences in identities, as encoded by $\boldsymbol\Sigma_\mathcal{I}=\mathbf{P}_\text{id}\boldsymbol\Sigma_\text{id}\mathbf{P}_\text{id}^\top$.
As expected, the shape uncertainties of expressions, quantified by $\boldsymbol\Sigma_\mathcal{E}$ are usually localized around the mouth and chin, as well as the regions around cheek and eyebrow.
More importantly, the variation by the residual term in Eq.~\eqref{eq:face_model_with_random_variables} has a much lower magnitude than those caused solely by identity and expression.
Note that the shape variations from identity or expression heavily depend on the dataset statistics that underpin this probabilistic multilinear model.

\subsubsection{Estimating Hyper-parameters}

We employ the FaceWarehouse dataset~\cite{cao2014facewarehouse} as the training dataset since it contains face meshes with a comprehensive set of expressions ($N_\text{exp}=50$) and a variety of identities ($N_\text{id}=150$) from different ages, genders and races.

Assigning each face mesh in the training dataset to two one-hot vectors $\mathbf{x}_\text{id}$ and $\mathbf{x}_\text{exp}$ for identity and expression, respectively, we find that $\mathbf{x}_\text{id}$/$\mathbf{x}_\text{exp}$ and $\mathbf{w}_\text{id}$/$\mathbf{w}_\text{exp}$ are linearly connected.
Because the face mesh is written as
$
\bar{\mathbf{f}} + \mathcal{T} \times_2 \mathbf{x}_\text{id}^\top \times_3 \mathbf{x}_\text{exp}^\top 
= \bar{\mathbf{f}} + \mathcal{C}\times_2 (\mathbf{U}_\text{id}^\top\mathbf{x}_\text{id})^\top \times_3 (\mathbf{U}_\text{exp}^\top\mathbf{x}_\text{exp})^\top
= \bar{\mathbf{f}} + \mathcal{C}\times_2 \mathbf{w}_\text{id}^\top \times_3 \mathbf{w}_\text{exp}^\top, \label{eq:face_model_representation}
$
we have $\mathbf{w}_\text{id} = \mathbf{U}_\text{id}^\top\mathbf{x}_\text{id}$ and $\mathbf{w}_\text{exp} = \mathbf{U}_\text{exp}^\top\mathbf{x}_\text{exp}$.

The mean face $\bar{\mathbf{f}}$ requires $\bar{\mathbf{x}}_\text{id} = \frac{1}{N_\text{id}}\mathbf{1}$ and $\bar{\mathbf{x}}_\text{exp} = \frac{1}{N_\text{exp}}\mathbf{1}$.
The variances $\text{Var}(\mathbf{x}_\text{id}) \simeq \frac{1}{N_\text{id}}\mathbf{I}$ and $\text{Var}(\mathbf{x}_\text{exp}) \simeq \frac{1}{N_\text{exp}}\mathbf{I}$, where $\mathbf{I}$ is the identity matrix.
Thus the hyper-parameters in the prior distributions can be estimated accordingly, such that $\boldsymbol\mu_\text{id} = \frac{1}{N_\text{id}}\mathbf{U}_\text{id}^\top \mathbf{1}$ and $\boldsymbol\mu_\text{exp} = \frac{1}{N_\text{exp}}\mathbf{U}_\text{exp}^\top \mathbf{1}$, and $\boldsymbol\Sigma_\text{id} \simeq \frac{1}{N_\text{id}}\mathbf{U}_\text{id}^\top \mathbf{U}_\text{id} =  \frac{1}{N_\text{id}} \mathbf{I}$ and $\boldsymbol\Sigma_\text{exp} \simeq \frac{1}{N_\text{exp}}\mathbf{U}_\text{exp}^\top \mathbf{U}_\text{exp} = \frac{1}{N_\text{exp}}\mathbf{I}$.

\section{Probabilistic Facial Pose Tracking}
\label{sec:probabilistic_facial_pose_tracking}

\begin{figure}[t]
\centering
\includegraphics[width=1\linewidth]{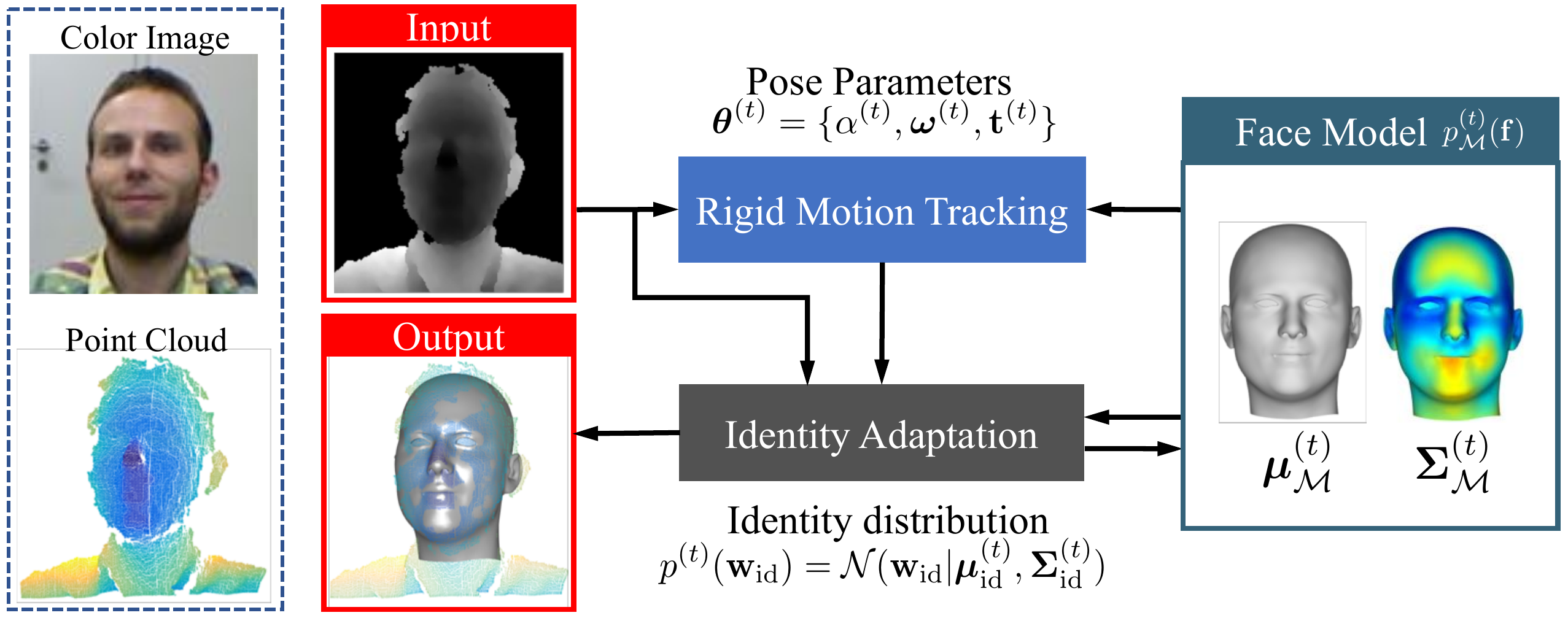}
\caption{Overview of the propose probabilistic framework, which consists of 1) robust facial motion tracking and 2) online switchable identity adaptation.
For both components, the generative model $p_\mathcal{M}^{(t)}(\mathbf{f})$ acts as the key intermediate and it is updated immediately with the feedback of the identity adaptation.
The input to the system is the depth map while the output is the rigid pose parameter $\boldsymbol\theta^{(t)}$ and the updated face identity parameters $\{\mu_\text{id}^{(t)}, \boldsymbol\Sigma_\text{id}^{(t)}\}$ that encode the identity distribution $p^{(t)}(\mathbf{w}_\text{id})$. Note that the color image is for illustration but not used in our system.}
\label{fig:system_overview}
\end{figure}

In this section, we present our probabilistic facial pose tracking approach.
Fig.~\ref{fig:system_overview} shows the overall architecture, which consists of two main components: 1) robust facial pose tracking, and 2) online switchable identity adaptation.
The goal of the first component is to estimate the rigid facial pose $\boldsymbol\theta$, given an input depth image and the current facial model.
The second component aims to update the distribution of the identity parameter $\mathbf{w}_\text{id}$ and the face model $p_\mathcal{M}(\mathbf{f})$, given the previous face model, the current pose parameter, and the input depth image.

\subsection{Notation and Prerequisites}
\label{sub:prerequisites}

\noindent\textbf{Notation.}
The input depth sequence is $\{ \mathbf{D}_t \}_{t=1}^T$, where $T$ is the number of frames.
The goal of the proposed system is to estimate the facial poses as $\{ \boldsymbol\theta^{(t)} \}_{t=1}^T$, and estimate all identities $\{ \mathcal{I}_k \}_{k=1}^K$ contained in this sequence.
Each identity $\mathcal{I}_k$ is parameterized by identity parameters in each identity distribution, please refer to Sec.~\ref{sub:online_switchable_identity_adaptation} for detailed description.

\vspace{+1mm}
\noindent\textbf{2D and 3D Conversion.}
Let the matrix-form camera parameters be $\mathbf{K} = \left[\begin{smallmatrix}
f & 0 & u_\mathbf{o} \\0 & f & v_\mathbf{o} \\ 0 & 0 & 1
\end{smallmatrix}\right]$, where $f$ is the focal length and $\mathbf{o} = [u_\mathbf{o}, v_\mathbf{o}]$ is the principle point.
A 3D point $\mathbf{p} = [x, y, z]^\top$ can be perspectively projected onto the pixel coordinate
\begin{equation}
[u, v, 1]^\top = \pi(\mathbf{p}) = \mathbf{K}[x/z, y/z, 1]^\top
\end{equation}
Inversely, a pixel $\mathbf{x} = [u, v]$ can be back-projected as 
\begin{equation}
\mathbf{p} = \pi^{-1}(\mathbf{x}, \mathbf{D}(\mathbf{x})) = \mathbf{K}^{-1} [u, v, 1]^\top \cdot \mathbf{D}(\mathbf{x}), \label{eq:inverse_projection}
\end{equation}
where $\mathbf{D}(\mathbf{x})$ is the depth at the pixel $\mathbf{x}$.

\vspace{+1mm}
\noindent\textbf{Rigid Transformation.}
The rigid facial pose $\boldsymbol\theta$ consists of the rotation angles $\boldsymbol\omega \in \mathbb{R}^3$, the translation vector $\mathbf{t} \in \mathbb{R}^3$, and an auxiliary scale factor $\alpha$.
Thus $\boldsymbol\theta = \{\boldsymbol\omega, \mathbf{t}, \alpha\}$ indicates a transformation of the face model,
\begin{equation}
\mathbf{q}_n = \mathbf{T}(\boldsymbol\theta) \circ \mathbf{f}_n = e^\alpha \mathbf{R}(\boldsymbol\omega)\mathbf{f}_n + \mathbf{t}, n \in \{1, \ldots, N_\mathcal{M}\} \label{eq:rigid_transform}
\end{equation}
where $\mathbf{R}(\boldsymbol\omega) \in \mathbb{R}^{3\times3}$ is the rotation matrix converted from $\boldsymbol\omega$ by Rodrigues' rotation formula.
The exponential scale $e^\alpha$ ensures the necessary positivity of the scale factor, $\forall\alpha\in\mathbb{R}$.

The auxiliary scale factor is introduced to model possible deviations of scale beyond that observed in the training dataset, for example when tracking children.
Even though iterative optimization alternating the model personalization and rigid pose estimation could alone resolve this issue, an explicit global scale will effectively compensate the scale incompatibility and speed up the optimization.

\vspace{+1mm}
\noindent\textbf{Transformed Face Distribution.}
We have depicted the probabilistic formulation for the morphable face model in Sec.~\ref{sec:probabilistic_3d_face_parameterization}.
The rigid transformed face model $\mathcal{Q}$ has a similar marginal distribution for each $\mathbf{q}_n\in\mathcal{Q}$ as Eq.~\eqref{eq:face_model_marginalization}, but with rotation:
\begin{equation}
p_\mathcal{Q}(\mathbf{q}_n;\boldsymbol\theta) = \mathcal{N}(\mathbf{q}_n|\mathbf{T}(\boldsymbol\theta)\circ \boldsymbol\mu_{\mathcal{M}, [n]}, e^{2\alpha}\boldsymbol\Sigma_{\mathcal{M},[n]}^{(\boldsymbol\omega)}), 
\label{eq:face_model_distribution}
\end{equation}
where $\boldsymbol\mu_{\mathcal{M},[n]}$ and $\boldsymbol\Sigma_{\mathcal{M},[n]}^{(\boldsymbol\omega)}$ are the mean and the rotated variance matrix for point $\mathbf{f}_n$, respectively.
Moreover, we have $\boldsymbol\Sigma_{\mathcal{M},[n]}^{(\boldsymbol\omega)} = \mathbf{R}(\boldsymbol\omega)\boldsymbol\Sigma_{\mathcal{M},[n]}\mathbf{R}(\boldsymbol\omega)^\top$.
$\boldsymbol\mu_{\mathcal{M}, [n]}$ and $\boldsymbol\Sigma_{\mathcal{M}, [n]}$ are the $n^\text{th}$ blocks corresponding to $\mathbf{f}_n$ in $\boldsymbol\mu_{\mathcal{M}}$ and $\boldsymbol\Sigma_{\mathcal{M}}$.

\subsection{Face Localization}
\label{sub:face_localization}

The face localization procedure infers the probable area containing a face in a depth image.
It is a vital preprocessing step to extract the point cloud $\mathcal{P}$ from the depth image and a fairly good pose initialization for the face tracking system.

\vspace{+1mm}
\noindent\textbf{System Initialization.}
By adopting an efficient filtering-based head localization method~\cite{meyer2015robust} to the first frame $\mathbf{D}_1$, the face ROI is localized at the pixel $\mathbf{c}$ in which the highest correlation is achieved with the average human head-shoulder template.
The size of the ROI is depth-adaptive, where the height $h = f\bar{h}/d_\mathbf{c}$ and the width is $w = f\bar{w}/d_\mathbf{c}$.
$d_\mathbf{c} = \mathbf{D}_1(\mathbf{c})$.
$\bar{h} = 240 \text{mm}$ is empirically set as the average head height, and $\bar{w} = 320 \text{mm}$ is around twice the size of the average head width, which generally ensures the coverage of the entire faces.
The point cloud $\mathcal{P}$ is a set of 3D points converted from depth pixels in the face ROI.
In addition, the initial pose consists of no rotation as $\boldsymbol\omega = \mathbf{0}$, and a translation $\mathbf{t} = \pi^{-1}(\mathbf{c}, d_\mathbf{c})$. Initially $\alpha = 0$ in default.

\vspace{+1mm}
\noindent\textbf{During Tracking.}
The point cloud is extracted in a manner similar to the system initialization, but the center pixel from the previous frame is corrected by the translation vector.
The pose is initialized from the previous frame pose estimate.

\subsection{Robust Facial Pose Tracking}

\begin{figure}[t]
\centering
\includegraphics[width=\linewidth]{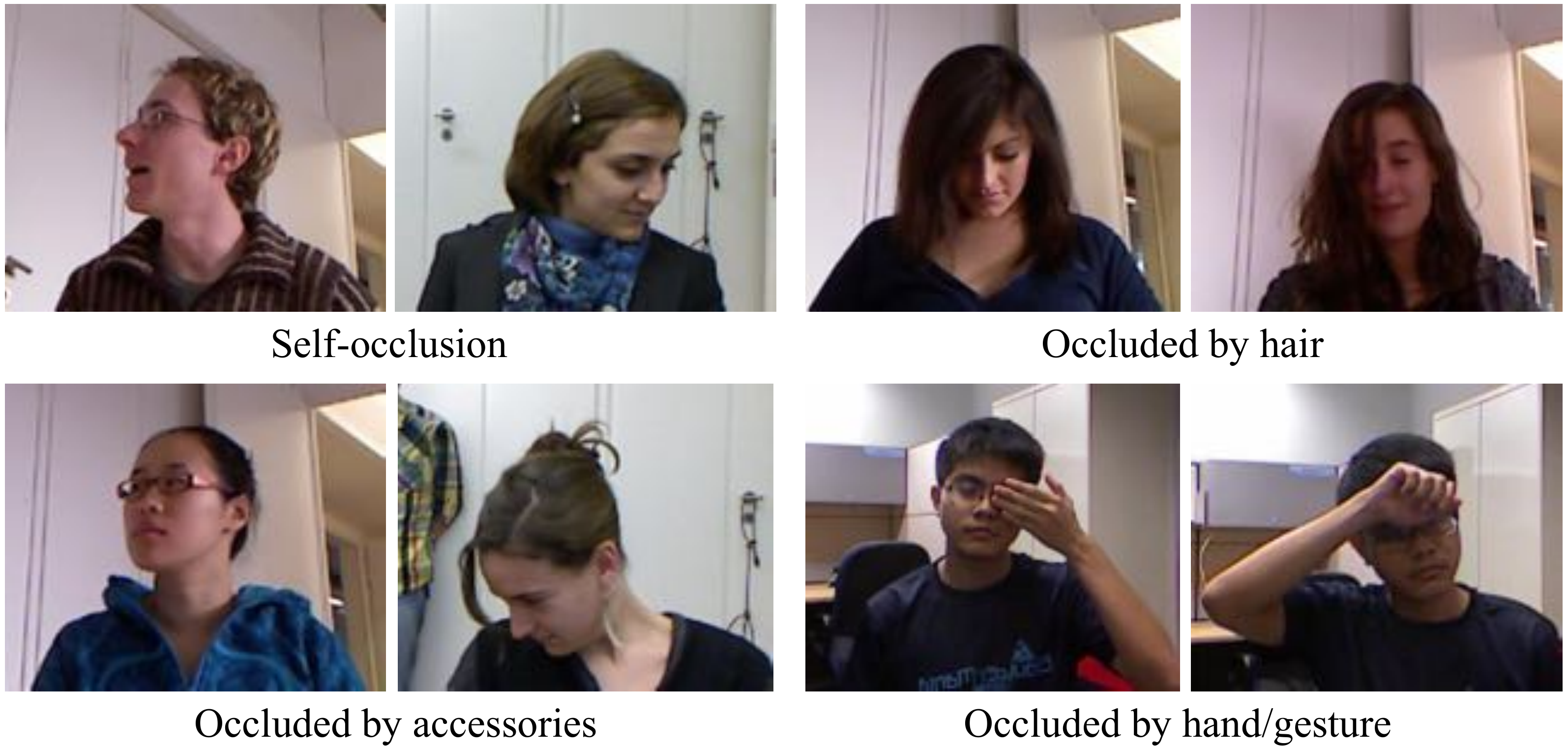}
\caption{Samples of the occluded faces.}
\label{fig:face_occlusion}
\end{figure}

An optimal pose suggests that the input point cloud $\mathcal{P}$ should fail within a high density region of the distribution of the warped face model $\mathcal{Q}$.
However, in uncontrolled scenarios, we often encounter self-occlusions or object-to-face occlusions, like hair, glasses and fingers/hands, as shown in Fig.~\ref{fig:face_occlusion}.
In these scenarios, even if the face model $\mathcal{Q}$ and the point cloud $\mathcal{P}$ are correctly aligned, $\mathcal{Q}$ can only partially fit a subset of 3D points in $\mathcal{P}$ while leaving the remaining points in $\mathcal{Q}$ occluded.

Therefore, it is important to find the visible parts of $\mathcal{Q}$, based on which we can robustly track the facial pose.
We do not follow a correspondence-based methods like distance thresholding and normal vector compatibility check~\cite{hsieh2015unconstrained} to identify the visible regions, since finding reliable correspondences is itself challenging.
Instead, we propose a \emph{ray visibility constraint} to regularize the visibility of each face model point, based on our developed statistical face prior.

\subsubsection{Ray Visibility Constraint}
\label{ssub:ray_visibility_constraint}

Denote the ray connecting the camera center to a face model point $\mathbf{q}_n$ as $\vec{v}_{\mathbf{q}_n}$.
This ray intersects with the point cloud $\mathcal{P}$ at a point $\mathbf{p}_n$, which can be found by matching the pixel location of $\mathbf{q}_n$ in the input depth image~\cite{hsieh2015unconstrained,li2013realtime}.

\begin{figure}[t]
\centering
\includegraphics[width=\linewidth]{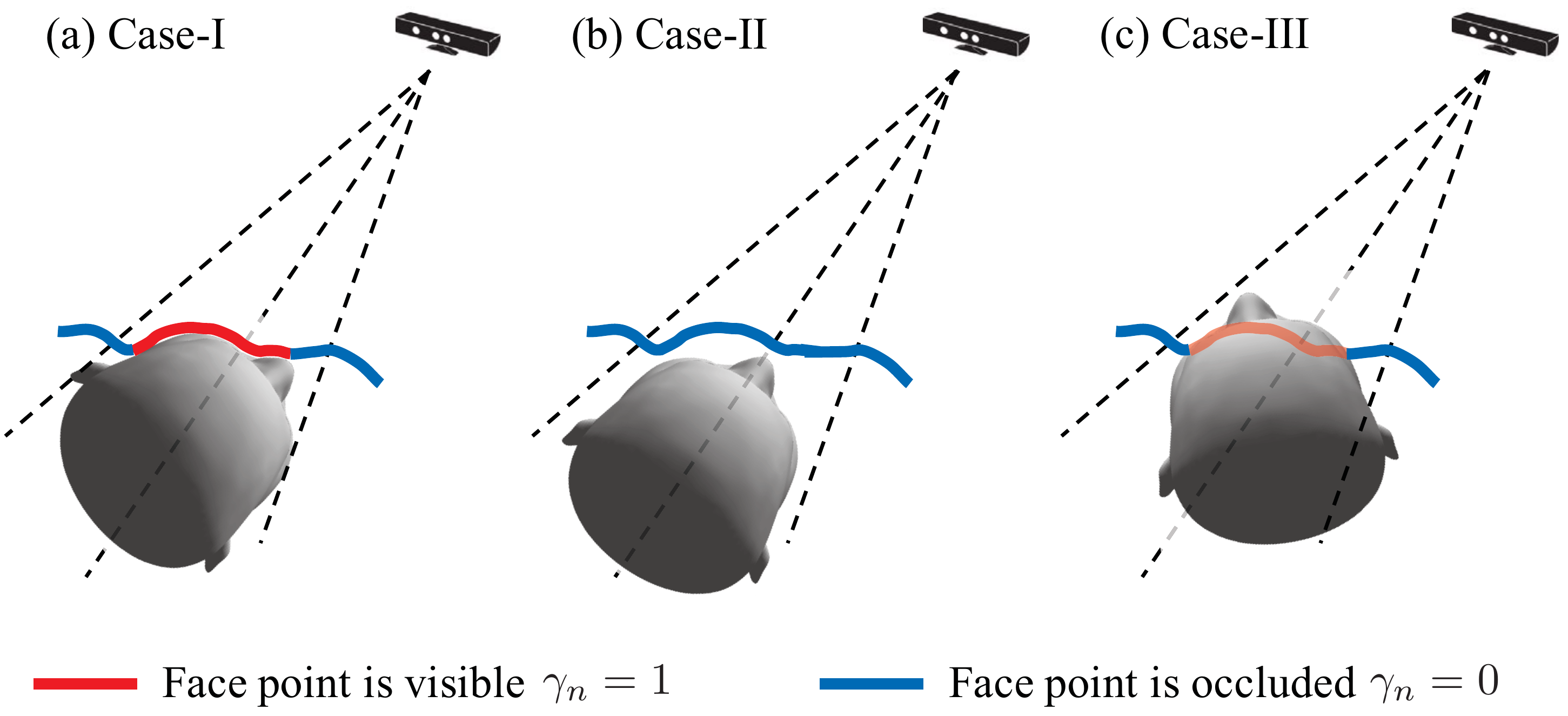}
\caption{Illustration of the ray visibility constraint. A profiled face model and a curve on the point cloud are in front of a depth camera. (a) A part of face points fit the curve, while the rest points are occluded. (b) The face model is completely occluded. (c) An infeasible case where the face model occludes the point cloud curve.}
\label{fig:ray_visibility_constraint}
\end{figure}

The role of the proposed ray visibility constraint (RVC) is to examine the physically reasonable relative position between $\mathbf{p}_n$ and $\mathbf{q}_n$ if the face model complies with a valid facial pose:
\begin{enumerate}
\item \emph{Visible} $\mathbf{q}_n$ is close to the local surface around the connected $\mathbf{p}_n$ in the point cloud $\mathcal{P}$;
\item \emph{Occluded} $\mathbf{q}_n$ must be located further away and behind the surface around $\mathbf{p}_n$;
\item \emph{Invalid} $\mathbf{q}_n$ is in front of the surface around $\mathbf{p}_n$, which is impossible and should be avoided.
\end{enumerate}
Take Fig.~\ref{fig:ray_visibility_constraint} as an example.
In this study, we propose a statistical formulation for the RVC constraint to automatically adjust the pose of the face model $\mathcal{Q}$.
Eventually, the face model will tightly but partially fit the point cloud $\mathcal{P}$ while leaving the rest of the points as occlusions.
The shape uncertainty by expressions are also absorbed in this constraint, so that the expression variations will not harm the pose estimation. 
Note that in our setting, \emph{invalid} point pairs usually suffer obligatory penalties to push the invalid face points farther away.

\begin{figure}[t]
\centering
\includegraphics[width=\linewidth]{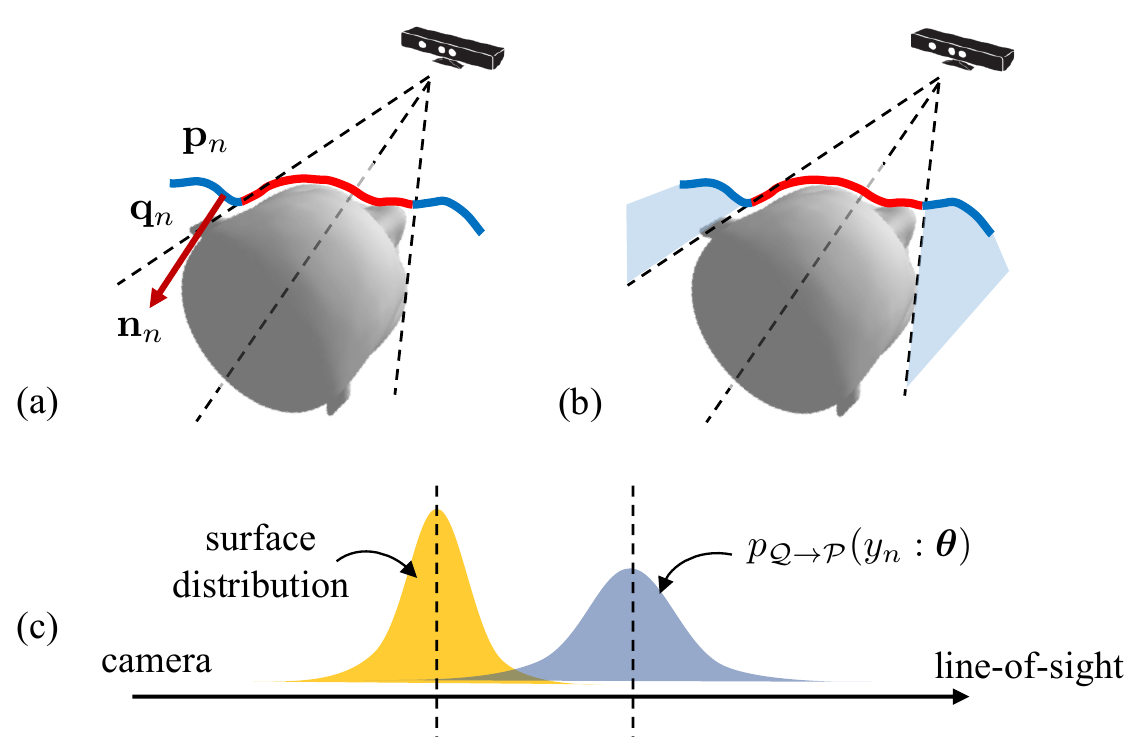}
\caption{(a) The signed distance through a ray $\vec{v}_{\mathbf{q}_n}$, where $\mathbf{n}_n$ is the normal vector at the surface point $\mathbf{p}_n$ that intersects $\vec{v}_{\mathbf{q}_n}$. (b) Visibility of the face model with respect to the point cloud. The occlusion space is marked in blue, while the visible points are marked in red. (c) Illustration of the surface distribution $\mathcal{N}(y_n| 0, \sigma_o^2)$ and the projected face distribution $p_{\mathcal{Q}\rightarrow\mathcal{P}}(y_n; \boldsymbol\theta)$, where the overlapping interval in between illustrates whether the face model is visible.}
\label{fig:illustration_distributions}
\end{figure}

\vspace{+1mm}
\noindent\textbf{Statistical Formulation.}
Assume the surface of $\mathcal{P}$ is $y = \mathbf{n}_n^\top (\mathbf{p} - \mathbf{p}_n)$, where $\mathbf{n}$ is the normal vector at $\mathbf{p}_n$ and $y$ is the signed distance of $\mathbf{p}$ onto this plane.
Thus the signed distance $y_n$ of a face point $\mathbf{q}_n$ to the surface around $\mathbf{p}_n$ is 
\begin{equation}
y_n = \Delta(\mathbf{q}_n; \mathbf{p}_n) = \mathbf{n}_n^\top (\mathbf{q}_n - \mathbf{p}_n),
\end{equation}
as visualized in Fig.~\ref{fig:illustration_distributions}(a).
Similarly as Eq.~\eqref{eq:face_model_distribution}, the distribution of the signed distance $p_{\mathcal{Q}\rightarrow\mathcal{P}}(y_n; \boldsymbol\theta)$, called the projected face distribution onto the surface around $\mathcal{P}$, will be
\begin{equation}
\mathcal{N}\left(y_n | \Delta(\mathbf{T}(\boldsymbol\theta)\circ\boldsymbol\mu_{\mathcal{M},[n]};\mathbf{p}_n), \sigma_o^2 + e^{2\alpha}\mathbf{n}_n^\top \boldsymbol\Sigma_{\mathcal{M},[n]}^{(\boldsymbol\omega)}\mathbf{n}_n\right), \label{eq:signed_distance_distribution}
\end{equation}
where $\sigma_o^2$ is the noise variance describing the surface modeling errors and the depth sensor's systematic errors.
Moreover, the surface distribution around $\mathbf{p}_n$ is assumed Gaussian $p_{\mathcal{P}}(y_n) = \mathcal{N}(y_n|0, \sigma_o^2)$, whose variance subsumes the modeling and systematic noise.

By profiling the distributions of the point pair $\{\mathbf{p}_n, \mathbf{q}_n\}$ from the ray $\vec{v}_{\mathbf{q}_n}$ in Fig.~\ref{fig:illustration_distributions}(c), the visibility (as $\gamma_n = 1$) is interpreted as the projected face distribution $p_{\mathcal{Q}\rightarrow\mathcal{P}}(y_n;\boldsymbol\theta)$ ``overlapping'' or being ``in front'' of the surface distribution $p_{\mathcal{P}}(y_n)$.
More precisely, we define visible surface points $\mathbf{q}_n$ as those for which $\Delta(\mathbf{T}(\boldsymbol\theta)\circ\boldsymbol\mu_{\mathcal{M},[n]};\mathbf{p}_n)$ is inside one standard deviation significance of $p_{\mathcal{Q}\rightarrow\mathcal{P}}(y_n)$ or negative\footnote{The normal $\mathbf{n}_n$ is set to point away from the camera center, thus \emph{negative} $y_n$ means that the face point $\mathbf{q}_n$ is in front of the surface of $\mathbf{p}_n$.},
\begin{equation}
\gamma_n=1: ~~\Delta(\mathbf{T}(\boldsymbol\theta)\circ\boldsymbol\mu_{\mathcal{M},[n]};\mathbf{p}_n) \leq \sqrt{\sigma_o^2 + e^{2\alpha}\mathbf{n}_n^\top \boldsymbol\Sigma_{\mathcal{M},[n]}^{(\boldsymbol\omega)} \mathbf{n}_n}. \notag
\end{equation}
The occlusion (as $\gamma_n = 0$) requires that $p_{\mathcal{Q}\rightarrow\mathcal{P}}(y_n;\boldsymbol\theta)$ is ``behind'' the major mass of the surface distribution, thus the signed distance $y_n$ should always be positive and beyond the confidence interval of $p_{\mathcal{Q}\rightarrow\mathcal{P}}(y_n; \boldsymbol\theta)$, \ie,
\begin{equation}
\gamma_n=0: ~~\Delta(\mathbf{T}(\boldsymbol\theta)\circ\boldsymbol\mu_{\mathcal{M},[n]};\mathbf{p}_n) > \sqrt{\sigma_o^2 + e^{2\alpha}\mathbf{n}_n^\top \boldsymbol\Sigma_{\mathcal{M},[n]}^{(\boldsymbol\omega)} \mathbf{n}_n}. \notag
\end{equation}

\subsubsection{Ray Visibility Score}
\label{ssub:ray_visibility_score}

The ray visibility score (RVS) is converted from the RVC for the rigid facial pose estimation, which measures the compatibility between the distributions of the transformed face model $\mathcal{Q}$ and the input point cloud $\mathcal{P}$.

For a ray $\vec{v}_{\mathbf{q}_n}$ between the face model point $\mathbf{q}_n$ and the cloud point $\mathbf{p}_n$, as the visibility $\gamma_n$ for $\mathbf{q}_n$ is judged by the pose $\boldsymbol\theta$, the distribution of $\mathbf{p}_n$ is now refined to be as
\begin{equation}
p_\mathcal{P}(y_n; \boldsymbol\theta) = \mathcal{N}(y_n|0,\sigma_o^2)^{\gamma_n} \mathcal{U_O}(y_n)^{1-\gamma_n}, \label{eq:point_cloud_distribution}
\end{equation}
where $\mathcal{U_O}(y_n) = U_\mathcal{O}$ is a uniform distribution that is valid within a range when $0 < y_n<2500$mm.
\eqref{eq:point_cloud_distribution} takes into account the visibility labels.
When $\mathbf{q}_n$ is visible, $\mathbf{p}_n$ has a compatible surface distribution of $\mathcal{N}(y_n|0,\sigma_o^2)$.
However, if $\mathbf{q}_n$ is occluded, $\mathbf{p}_n$ can be arbitrary as long as it is in front of $\mathbf{q}_n$, which we model as a uniform distribution $\mathcal{U_O}(y_n)$.
$p_\mathcal{P}(y_n; \boldsymbol\theta)$ can be regarded as a noisy face measurements contaminated by occlusions, while the projected face distribution $p_{\mathcal{Q}\rightarrow\mathcal{P}}(y_n;\boldsymbol\theta)$ represents the face model with its own uncertainties along the local surfaces.
If ideally aligned, these distributions should be co-located and cover each other in a statistical manner.

The RVS score $\mathcal{S}(\mathcal{Q}, \mathcal{P}; \boldsymbol\theta)$ is thus to measure the similarity between $p_\mathcal{P}(\mathbf{y}; \boldsymbol\theta) = \prod_{n=1}^{N_\mathcal{M}} p_\mathcal{P}(y_n; \boldsymbol\theta)$ and $p_\mathcal{Q}(\mathbf{y};\boldsymbol\theta) = \prod_{n=1}^{N_\mathcal{M}} p_{\mathcal{Q}\rightarrow\mathcal{P}}(y_n; \boldsymbol\theta)$ by the Kullback-Leibler divergence,
\begin{equation}
\mathcal{L}_\text{rvs}(\boldsymbol\theta) = D_{KL}\left[p_\mathcal{Q}(\mathbf{y};\boldsymbol\theta) || p_\mathcal{P}(\mathbf{y}; \boldsymbol\theta)\right] \label{eq:ray_visibility_score}
\end{equation}
so that the more similar $p_\mathcal{P}(\mathbf{y}; \boldsymbol\theta)$ and $p_\mathcal{Q}(\mathbf{y}; \boldsymbol\theta)$ are, the smaller $\mathcal{L}_\text{rvs}(\boldsymbol\theta)$ is.
Thus, the optimal pose parameter $\boldsymbol\theta^*$ is the one minimizing the RVS score $
\boldsymbol\theta^* = \arg \min_{\boldsymbol\theta} \mathcal{L}_\text{rvs}(\boldsymbol\theta)$.
The visibility labels are instantaneously obtained when the pose parameter is given.
Note that Eq.~\eqref{eq:ray_visibility_score} not only accounts for the visible points but also penalizes the number of occluded points, thus avoiding a degenerated solution where a majority of the face points are labeled as occluded.

\subsubsection{Temporal Constraint}
\label{ssub:temporal_enhancement}

The single-frame rigid pose estimation can effectively employ the proposed ray visibility score, however the lack of temporal cohesion results in unstable pose and identity estimation over a series of frames.
In this part, we try to relieve these drawbacks by enforcing temporal pose coherence over adjacent frames.
Since a fixed identity should bear a fixed scale, thus the estimated scale should be accumulated over a long period of sequence to stabilize the identity estimation.

\vspace{+1mm}
\noindent\textbf{Temporal Coherence.}
Observing that the face model and the point clouds should concurrently follow the same rigid motion between adjacent frames, it is possible to apply the pixel-wise flows in the input sequences to regularize the pose changes of the face model.
Consider a facial pixel $\mathbf{x}_m^{t-1}$ at frame $t-1$.
A change in pose of $\Delta\boldsymbol\theta$ will induce the change of the pixel location, such as:
\begin{equation}
\mathbf{W}(\mathbf{x}_m^{t-1}, \Delta\boldsymbol\theta) = \pi(\mathbf{T}(\Delta\boldsymbol\theta)\circ\pi^{-1}(\mathbf{x}_m^{t-1}, \mathbf{D}_{t-1}(\mathbf{x}_m^{t-1}))).
\end{equation}
Therefore, an optimal $\Delta\boldsymbol\theta$ will eliminate the depth difference between $\mathbf{D}_{t}(\mathbf{W}(\mathbf{x}_m^{t-1}, \Delta\boldsymbol\theta))$ in frame $t$ and the depth value of the transformed point as $\mathcal{Z}(\mathbf{x}_m^{t-1};\Delta\boldsymbol\theta) = [0, 0, 1]\mathbf{T}(\Delta\boldsymbol\theta)\circ\pi^{-1}(\mathbf{x}_m^{t-1}, \mathbf{D}_{t-1}(\mathbf{x}_m^{t-1}))$.
We use a temporal smoothness term for pose changes in adjacent frames to account for this fact, \ie,
\begin{equation}
\mathcal{L}_t(\Delta\boldsymbol\theta) = \frac{1}{2\sigma_t^2}\sum_{m=1}^{M^{t-1}} \left(\mathbf{D}_{t}(\mathbf{W}(\mathbf{x}_m^{t-1}, \Delta\boldsymbol\theta)) - \mathcal{Z}(\mathbf{x}_m^{t-1};\Delta\boldsymbol\theta) \right)^2,
\end{equation}
for a set of visible pixels $\{ \mathbf{x}_m^{t-1} \}_{m=1}^{M^{t-1}}$ in frame $t-1$, with the temporal variance controlled by $\sigma_t^2$.

In comparison with the prediction-correction techniques like Kalman filtering, which smooth the facial poses according to the statistics of their preceding estimates, the proposed method explicitly relates the flows between adjacent facial points and the motion of the face model, and is capable of rapidly capturing the incremental motion without pose flickers, and avoiding over-smoothing introduced by online filtering like Kalman filtering or average filtering~\cite{bishop2006pattern}.

\begin{algorithm}[t]
\begin{small}
\LinesNumbered
\SetKwInOut{Input}{Input}\SetKwInOut{Output}{Output}
\Input{Input depth frame $\mathbf{D}_t$\; \\
Previous pose parameters $\boldsymbol\theta^{(t-1)}$\; \\
Previous visible face pixels $\{\mathbf{x}_m^{t-1}\}_{m=1}^{M^{t-1}}$\;
}
\Output{Current pose parameters $\boldsymbol\theta^{(t)}$ \;\\
Current visible face pixels $\{\mathbf{x}_m^{t}\}_{m=1}^{M^{t}}$\; \\
}
\BlankLine
$\boldsymbol\theta_0 \gets \boldsymbol\theta^{(t-1)}$ or initialization by face localization from $\mathbf{D}_t$\;
$\mathcal{P}^{(t)} \gets $ localize face region from $\mathbf{D}_t$ based on $\boldsymbol\theta_0$\;
$\Delta\boldsymbol\theta \gets \mathbf{0}$\;
\While{$i < N_{iter}$ or $\Delta\boldsymbol\theta$ has not converged}
{
  $\{ \gamma_n^{(t)}\}_{n=1}^{N_\mathcal{M}} \gets $ visibility detection via ray visibility constraint from $\mathcal{P}^t$ and $\mathcal{Q}$ warped by $\boldsymbol\theta_0 + \Delta\boldsymbol\theta$\;
  $\Delta\boldsymbol\theta \gets $ quasi-Newton update of $\mathcal{L}(\Delta\boldsymbol\theta)$\;
  $i \gets i + 1$\;
}
Augment $\Delta\boldsymbol\theta$ by PSO when tracking failed\; 
$\boldsymbol\theta^{(t)} \gets \boldsymbol\theta_0 + \Delta\boldsymbol\theta$\;
Extract visible face pixels $\{\mathbf{x}_m^{t}\}_{m=1}^{M^t}$ by reprojection
\end{small}
\caption{Robust 3D Facial Pose Tracking}
\label{alg:robust_3d_facial_pose_tracking}
\end{algorithm}

\vspace{+1mm}
\noindent\textbf{Scale Accumulation.}
The scale $e^\alpha$ is indeed related to the face model rather than a component in the rigid pose, it means that a fixed tracked subject should have a fixed scale.
Therefore, the estimation of the scale factor requires explicit accumulations from the preceding estimates if the subject keeps the same. 
The scale loss accounts for the mismatch between the previous scale and the current estimate
\begin{equation}
\mathcal{L}_{s}(\Delta\boldsymbol\theta) = \frac{1}{2}\lambda^{(t)}_s(\Delta\alpha)^2,
\end{equation}
where $\Delta\alpha = \alpha - \alpha^{(t-1)}$, and $\alpha^{(t-1)}$ is the estimated scale factor in the previous frame.
$\lambda^{(t)}_s$ is the cumulative precision across the frames, written as $\lambda^{(t)}_s = \lambda^{(t-1)}_s + \frac{1}{\sigma_s^2}$, where $\sigma_s^2$ is a predefined variance of the scale factor.

\subsubsection{Rigid Pose Estimation}
\label{ssub:rigid_pose_estimation}

The overall formulation for the rigid pose tracking combines the ray visibility score and the temporal constraint, as 
\begin{equation}
\mathcal{L}(\Delta\boldsymbol\theta) = \mathcal{L}_\text{rvs}(\boldsymbol\theta^{(t-1)}+\Delta\boldsymbol\theta) + \mathcal{L}_t(\Delta\boldsymbol\theta) + \mathcal{L}_s(\Delta\boldsymbol\theta). \label{eq:rigid_pose_overall_formulation}
\end{equation}
We seek to estimate the incremental pose parameters $\Delta\boldsymbol\theta$ between adjacent frames rather than the absolute poses $\boldsymbol\theta$ between the canonical face model and the input point cloud.

Solving Eq.~\eqref{eq:rigid_pose_overall_formulation} is challenging since $\mathcal{L}_\text{rvs}(\boldsymbol\theta^{(t-1)}+\Delta\boldsymbol\theta)$ is highly nonlinear with no closed-form solution.
In this work, we apply a recursive estimation method.
In particular, in each iteration, we alternatively determine the visibility labels $\boldsymbol\gamma$ given the previous pose parameters, and then estimate the incremental pose parameters $\Delta\boldsymbol\theta$.
In the first sub-problem for visibility labels, we examine the ray visibility constraint to all point pairs along $\{\vec{v}_{\mathbf{q}_n}\}_{n=1}^{N_\mathcal{M}}$ coming from the current pose estimate.
In the second sub-problem for incremental pose estimation, we apply the quasi-Newton update using the trust region approach for the overall cost, given the current visibility labels.
The process repeats until convergence or beyond the predefined iteration numbers.

{
Two criteria are applied for detecting the bad local minima: 1) Estimated poses are unreasonable, \ie, sudden pose changes (\eg, large rotation changes $|\Delta\boldsymbol\omega| > \pi/4$ or large translation changes $|\Delta\mathbf{t}| > 100$ mm) over adjacent frames, and impossible pose parameters out of their possible ranges (\eg, as depicted in Tab.~\ref{tab:dataset_summary}). 2) The proportion of occluded pixels, relative to total number of pixels in the face region, exceeds 50\%.}
Therefore, particle swarm optimization (PSO)~\cite{padeleris2012head,meyer2015robust} is optionally added to tackle poor initialization and bad local minima.
Our facial pose tracking approach is listed in Algorithm~\ref{alg:robust_3d_facial_pose_tracking}, which highlights the key steps leading to the pose and visibility estimation.

\subsection{Online Switchable Identity Adaptation}
\label{sub:online_switchable_identity_adaptation}

Concurrently with the rigid pose tracking, the face model is also progressively adapted to the tracked subject's identity.
Moreover, the proposed identity adaptation can instantaneously switch to different users and instantiate novel face models for new identities.
To accomplish this, we track the switchable identity distributions in an online Bayesian updating scheme.

\begin{algorithm}[t]
\begin{small}
\LinesNumbered
\SetKwInOut{Input}{Input}\SetKwInOut{Output}{Output}
\Input{A depth video clip $\{\mathbf{D}_t\}_{t=1}^T$\; \\
Personalized identity models $\{ \mathcal{I}_k\}_{k=1}^{K}$\;
Presented identity model $\mathcal{I}_\text{present}$\;
}
\Output{Updated identity models $\{\mathcal{I}_k^\text{new}\}_{k=1}^{K_\text{new}}$\;\\
Updated presented identity model $\mathcal{I}_\text{present}^\text{new}$
}
\BlankLine
Generate the face model $\mathcal{Q}$ by $p_{\mathcal{I}_\text{present}}(\mathbf{w}_\text{id})$\;
$\mathcal{C}_k \gets \emptyset, \forall k \in \{0, \ldots, K\}$\;
\For{$t\gets1$ \textup{to} $T$}
{
  $\boldsymbol\theta^{(t)}, \{ \gamma_n^{(t)}\}_{n=1}^{N_\mathcal{M}} \gets$ rigid pose estimation for $\mathbf{D}_t$ with $\mathcal{Q}$\;
  $\kappa^{(t)} \gets \sum_{n=1}^{N_\mathcal{M}}\gamma_n^{(t)}$\;
  $\mathbf{w}_\text{id}^{(t)} \gets $ maximum likelihood estimation of Eq.~\eqref{eq:identity_likelihood}\;
  $k^{(t)} \gets \arg\max_k p(k|\mathbf{w}_\text{id}^{(t)}; \{\mathcal{I}\}_{k=1}^K, \mathcal{I}_0)$ \;
  $\mathcal{C}_{k^{(t)}} \gets \mathcal{C}_{k^{(t)}} \cup \{(\mathbf{w}_\text{id}^{(t)}, \kappa^{(t)})\}$\;
}
$\mathcal{I}_k^\text{new} \gets$ update by $\mathcal{C}_k$ via Eq.~\eqref{eq:m_and_beta_update}-\eqref{eq:nu_update}, $\forall k \in \{1, \ldots, K\}$\;
\If{$|\mathcal{C}_0| > 0$}
{
  $\mathcal{I}_{K+1} \gets \mathcal{I}_0$, $K_\text{new} \gets K + 1$\;
  $\mathcal{I}_{K+1}^\text{new} \gets $ update by $\mathcal{C}_0$ via Eq.~\eqref{eq:m_and_beta_update}-\eqref{eq:nu_update}\;
  \lIf{$k^{(T)} == 0$}{$k^{(T)}\gets K+1$}
}
$\mathcal{I}_\text{present}^\text{new} \gets \mathcal{I}^\text{new}_{k^{(T)}}$
\end{small}
\caption{Online Identity Adaptation}
\label{alg:online_identity_adaptation}
\end{algorithm}

\subsubsection{Online Identity Adaptation}
\label{ssub:online_identity_personalization}

As depicted in Sec.~\ref{sub:proposed_statistical_face_model}, the face model is personalized by the identity distribution $p^\star(\mathbf{w}_\text{id}) = \mathcal{N}(\mathbf{w}_\text{id}|\boldsymbol\mu_\text{id}^\star, \boldsymbol\Sigma_\text{id}^\star)$.
However, the exact $p^\star(\mathbf{w}_\text{id})$ is unknown without adequate depth samples.
Our goal is to sequentially update the face identity parameters $\boldsymbol\mu_\text{id}$ and $\boldsymbol\Sigma_\text{id}$ so as to gradually match the statistics of the input identity.

Formally, we apply an online Bayesian model to fulfill the sequential estimate.
The priors for $\boldsymbol\mu_\text{id}$ and $\boldsymbol\Sigma_\text{id}$ are assigned as Normal Inverse-Wishart conjugate priors~\cite{bishop2006pattern}:
\begin{equation}
p(\boldsymbol\mu_\text{id},\boldsymbol\Sigma_\text{id}) = \mathcal{N}(\boldsymbol\mu_\text{id}|\mathbf{m}, \beta^{-1}\boldsymbol\Sigma_\text{id}) \mathcal{W}^{-1}(\boldsymbol\Sigma_\text{id}|\boldsymbol\Psi, \nu)
\end{equation}
in which $\boldsymbol\Sigma_\text{id}$ follows the inverse-Wishart distribution $\mathcal{W}^{-1}$, and $\boldsymbol\mu_\text{id}$ conditionally depends on $\boldsymbol\Sigma_\text{id}$ and follows a Gaussian distribution.
Therefore, given a streaming set of samples $\mathbf{w}_\text{id}$, the parameters $\{\mathbf{m}, \beta, \boldsymbol\Psi, \nu \}$ can be analytically updated~\cite{bishop2006pattern}, and we may simply employ this set of parameters to estimate the expected mean $\mathbb{E}[\boldsymbol\mu_\text{id}] = \mathbf{m}$ and variance $\mathbb{E}[\mathbb{\boldsymbol\Sigma}_\text{id}] = \boldsymbol\Psi/(\nu - N_\text{id} - 1)$ of the identity distribution $p(\mathbf{w}_\text{id})$.

Therefore, the identity adaptation turns out to estimating the parameter set $\{ \mathbf{m}, \beta, \boldsymbol\Psi, \nu \}$ from a series of estimated $\{\mathbf{w}_\text{id}^{(t)}\}_{t=1}^T$.
And an additional identity parameter is the scale factor $\alpha$, estimated from the tracking parameters.
We thus denote the complete identity model as $\mathcal{I} = \{ \mathbf{m}, \beta, \boldsymbol\Psi, \nu, \alpha \}$.

{
\vspace{+1mm}
\noindent\textbf{-- Estimating $\mathbf{w}_\text{id}$.}
We directly estimate $\mathbf{w}_\text{id}$ by fitting the face model $\mathcal{Q}$ and the point cloud $\mathcal{P}$ in a maximum likelihood sense.
The likelihood consists of point-to-point and point-to-plane costs~\cite{rusinkiewicz2001efficient} for measuring the discrepancy between paired points in the visible region.
Thus, $\mathbf{w}_\text{id}^{(t)}$ in frame $t$ is achieved by maximizing
\begin{multline}
p_{\ell}(\mathbf{y}^{t}, \mathcal{P}^{t}|\mathbf{w}_\text{id};\boldsymbol\theta^{(t)}) = \prod_{n=1}^{N_\mathcal{M}}\mathcal{U}_\mathcal{O}(y_n^t)^{1-\gamma_n^{(t)}} \times \\
\prod_{n=1}^{N_\mathcal{M}} p_{\mathcal{Q}\rightarrow\mathcal{P}}(y_n^{t}|\mathbf{w}_\text{id}; \boldsymbol\theta^{(t)})^{\gamma_n^{(t)}} p_{\mathcal{Q}}(\mathbf{p}_n^{t}|\mathbf{w}_\text{id}; \boldsymbol\theta^{(t)})^{\gamma_n^{(t)}}, \label{eq:identity_likelihood}
\end{multline}
where $p_{\mathcal{Q}\rightarrow\mathcal{P}}(y_n^{t}|\mathbf{w}_\text{id}; \boldsymbol\theta^{(t)})$ and $p_{\mathcal{Q}}(\mathbf{p}_n^{t}|\mathbf{w}_\text{id}; \boldsymbol\theta^{(t)})$ are transferred from Eq.~\eqref{eq:face_model_distribution} and~\eqref{eq:signed_distance_distribution} but have the form of $p_\mathcal{M}(\mathbf{f}|\mathbf{w}_\text{id})$ in Eq.~\eqref{eq:face_model_identity}. 
The confidence of the estimated $\mathbf{w}_\text{id}^{(t)}$ is modeled by the portion of the visible region relative to the entire head area in the face model,
\begin{equation}
\kappa^{(t)} = \frac{1}{N_\mathcal{M}}\sum_{i=1}^{N_\mathcal{M}}\gamma_n^{(t)}. \label{eq:quality_factor}
\end{equation}
Therefore, given a clip of input frames, we can gather the identity parameters and their confidence scores as a tuple set $\mathcal{C} = \{(\mathbf{w}_\text{id}^{(1)}, \kappa^{(1)}), (\mathbf{w}_\text{id}^{(2)}, \kappa^{(2)}), \ldots, (\mathbf{w}_\text{id}^{(T)}, \kappa^{(T)}) \}$.

\vspace{+1mm}
\noindent\textbf{-- Updating $\{ \mathbf{m}, \beta, \boldsymbol\Psi, \nu \}$.}
Given the estimated identity parameters $\{ \mathbf{w}_\text{id}^{(t)} \}_{t=1}^T$, the distribution $p(\boldsymbol\mu_\text{id}, \boldsymbol\Sigma_\text{id})$ is updated via online parameter updating based on Bayesian posteriors
\begin{align}
& \mathbf{m}_\text{new} = \frac{N_\mathcal{C}\bar{\mathbf{w}}_\text{id}+\beta\mathbf{m}}{N_\mathcal{C}+\beta},~\text{and}~\beta_\text{new} = \beta + N_\mathcal{C} \label{eq:m_and_beta_update}\\
& \boldsymbol\Psi_\text{new} = \boldsymbol\Psi + N_\mathcal{C}\mathbf{S} + \frac{\beta N_\mathcal{C}}{\beta + N_\mathcal{C}}(\bar{\mathbf{w}}_\text{id}-\mathbf{m})(\bar{\mathbf{w}}_\text{id}-\mathbf{m})^\top, \label{eq:psi_update}\\
& \text{and}~\nu_\text{new} = \nu + N_\mathcal{C}, \label{eq:nu_update}
\end{align}
where $N_\mathcal{C} = \sum_{t=1}^T \kappa^{(t)}$, $\bar{\mathbf{w}}_\text{id} = \frac{1}{N_\mathcal{C}}\sum_{t=1}^T \kappa^{(t)} \mathbf{w}_\text{id}^{(t)}$ and $\mathbf{S} = \frac{1}{N_\mathcal{C}} \sum_{t=1}^{T} \kappa^{(t)} (\mathbf{w}_\text{id}^{(t)} - \bar{\mathbf{w}}_\text{id})(\mathbf{w}_\text{id}^{(t)} - \bar{\mathbf{w}}_\text{id})^\top$ as the weighted scatter matrix of the samples.
As more frames are acquired, the identity distribution more closely captures the statistics of the identity samples, where the mean value converges to the weighted mean of all samples, and its variance becomes narrower and converges to the samples' scatter matrix.
}

\subsubsection{Switchable Identity Adaptation}
\label{ssub:switchable_identity_adaptation}

To seamlessly create a novel identity as well as switch among multiple identities, we extend our method to an online mixture model.
The switching function is the posterior with respect to the identity ID $k$ conditioned on the current identity parameter $\mathbf{w}_\text{id}$ and the stored identity models $\{\mathcal{I}_k\}_{k=1}^K$:
\begin{equation}
p(k|\mathbf{w}_\text{id}; \{\mathcal{I}_k\}_{k=1}^K, \mathcal{I}_0) = \frac{p_{\mathcal{I}_k}(\mathbf{w}_\text{id})}{\sum_{k=1}^K p_{\mathcal{I}_k}(\mathbf{w}_\text{id}) + p_{\mathcal{I}_0}(\mathbf{w}_\text{id})}, \label{eq:switch_posterior}
\end{equation}
where $\mathcal{I}_0$ represents the generic identity model.
$p_{\mathcal{I}_k}(\mathbf{w}_\text{id})$ is the identity prior where $\boldsymbol\mu_\text{id}$ and $\boldsymbol\Sigma_\text{id}$ are controlled by $\mathcal{I}_k$.
This posterior assigns a sample $\mathbf{w}_\text{id}$ to one identity model $k$ according to the Gaussian mixture model.

Therefore, we can automatically cluster the set of identity samples and confidence scores $\mathcal{C} = \{ \mathbf{w}_\text{id}^{(t)}, \kappa^{(t)} \}_{t=1}^T$ into $\{\mathcal{C}_k\}_{k=1}^K$ according to this MAP estimation, and update the $k^\text{th}$ identity model with its corresponded clustered tuples.
{
Note that if the indicator once refers to the generic model $\mathcal{I}_0$, or the set $|\mathcal{C}_0|>0$, we instantiate a new identity model, initialized by the generic model, which is subsequently updated, as shown in lines 10 to 13 in Alg.~\ref{alg:online_identity_adaptation}.
}
The presented face model is assigned by the indicated identity model in the last frame, and its scale factor is parsed from the stored model accordingly.

\vspace{+1mm}
We summarize our online switchable identity adaptation in Alg.~\ref{alg:online_identity_adaptation}.
In practice, each identity continues adaptation until its adapted face model converges, \ie, the average point-wise mean squared error between adjacent face models is smaller than a given threshold.

\begin{figure}[t]
\includegraphics[width=\linewidth]{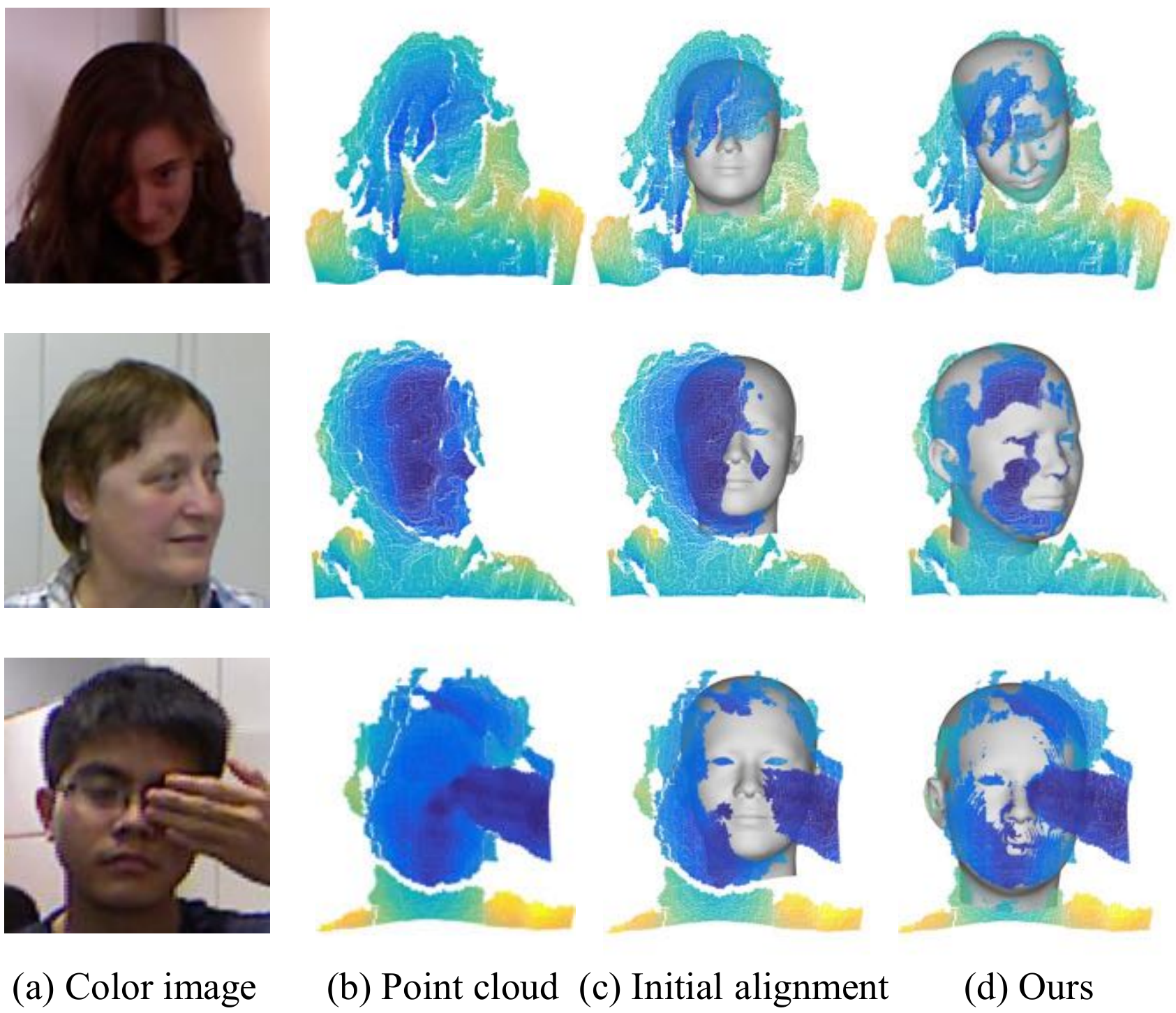}
\caption{Examples of our rigid pose estimation by the generic face model. (a)--(b): color images and the corresponded point clouds. (c): initial alignment provided by the proposed face localization method. (d): the proposed rigid pose estimation results.}
\label{fig:rigid_pose_estimation_examples}
\end{figure}

\begin{table}[t]
\centering
\caption{Facial Pose Datasets Summary}
\label{tab:dataset_summary}
\vspace{-3mm}
\begin{tabular}{c|c|c|c|c}
\hline
\hline
& $N_\text{seq}$ & $N_\text{frm}$ & $N_\text{subj}$ & $ \boldsymbol\omega_{\max}$ \\
\hline
\multirow{2}{*}{BIWI~\cite{fanelli2011real}} & \multirow{2}{*}{24} & \multirow{2}{*}{$\sim$15K} & \multirow{2}{*}{25} & $\pm 75^\circ$~yaw\\
 & & & & $\pm 60^\circ$~pitch \\
\hline
\multirow{2}{*}{ICT-3DHP~\cite{baltruvsaitis20123d}} & \multirow{2}{*}{10} & \multirow{2}{*}{$\sim$14K} & \multirow{2}{*}{10} & $\pm 75^\circ$~yaw\\
 & & & & $\pm 45^\circ$~pitch \\
\hline
\hline
\end{tabular}
\vspace{-3mm}
\end{table}

\section{Experiments and Discussions}

\subsection{Datasets and System Setup}
\label{ssec:dataset_and_setup}

\noindent\textbf{Datasets.}
We evaluate the proposed method on two public depth-based benchmark datasets, \ie, the Biwi Kinect head pose dataset~\cite{fanelli2011real} and ICT 3D head pose (ICT-3DHP) dataset~\cite{baltruvsaitis20123d}.
The dataset summaries are listed in Tab.~\ref{tab:dataset_summary}.

\textit{Biwi Dataset:}
$15$K RGB-D images of $20$ subjects (different genders and races) in $24$ sequences, with large ranges in rotations and translations. 
The recorded faces suffer the occlusions from hair and accessories and shape variations from facial expressions.
The groundtruth poses and face models were generated by RGB-D based FaceShift~\cite{weise2011realtime}.

\textit{ICT-3DHP Dataset:} $10$ Kinect RGB-D sequences including $6$ males and $4$ females.
This dataset contains similar occlusions and distortions like Biwi dataset. 
Each subject also involves arbitrary expression variations.
The groundtruth rotations were collected by Polhemus Fastrak flock of birds tracker that is attached to a cap worn by each subject.

\vspace{+1mm}
\noindent\textbf{Implementation.}
We implemented the proposed depth-based 3D facial pose tracking model in \texttt{MATLAB}.
The results reported in this paper were measured on a 3.4 GHz Intel Core i7 processor with 16GB RAM.
No GPU acceleration was applied.
The face model is specified as $N_\mathcal{M} = 11510, N_\text{id} = 150, N_\text{exp} = 47$.
In practice, we employ a truncated multilinear model with smaller dimensions as $\tilde{N}_\text{id} = 28, \tilde{N}_\text{exp} = 7$ for the sake of efficiency.
We set the noise variance as $\sigma_o^2 = 25$, and the outlier distribution is $\mathcal{U_O}(y) = U_\mathcal{O} = \frac{1}{2500}$.
$\sigma_t^2$ is empirically set to $75$, and $\sigma_s^2$ is $0.04$.
Note that the measurement unit in this paper is millimeter (mm).
The online face adaptation is performed every $5$ frames to avoid overfitting to partial facial scans.

\vspace{+1mm}
\noindent\textbf{Visualization.}
We plot the point clouds by \texttt{MATLAB}'s \emph{parula} color map.
Each point's color represents its depth to the camera, where the farther point has a warmer color, and the nearer point has a cooler color.
The presented face model is visualized as transformed face mesh based on Phong shading.
The visibility masks are marked in red and overlaid on the transformed face mesh.
{
Note that the face model has holes inside the mouth and eyes, and thus the results shown in Fig.~\ref{fig:poses_with_visibility_check}, Fig.~\ref{fig:exemplar_online_identity_adaptation}(c) and Fig.~\ref{fig:identity_reconstruction} are not meaningful.
}

\begin{figure}
\includegraphics[width=\linewidth]{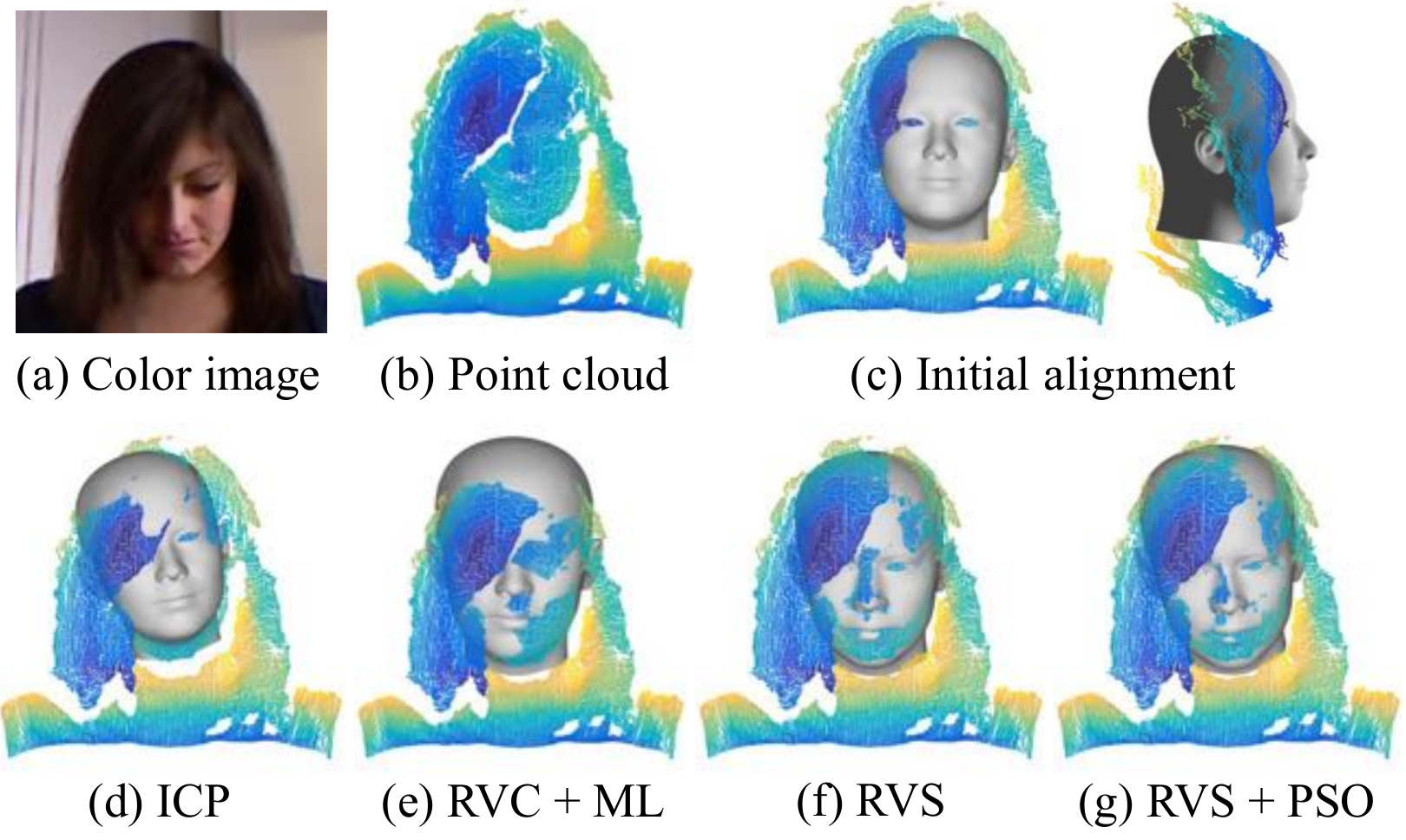}
\caption{Comparison of the rigid pose estimation methods with the generic face model. (a) and (b): color image and its corresponded point cloud. (c): two views of the initial alignment provided by face localization. (d): result by iterative closest points (ICP)~\cite{rusinkiewicz2001efficient}. (e): result by maximizing the log-likeligood $\log p_{\mathcal{Q}\rightarrow\mathcal{P}}(\mathbf{y};\boldsymbol\theta)$. (f): result by minimizing the ray visibility score. (g): the augmented RVS method by particle swarm optimization.}
\label{fig:rigid_pose_estimation_example_comparison}
\end{figure}

\begin{figure*}
\centering
\subfloat[Biwi dataset]{\includegraphics[height=0.30\linewidth]{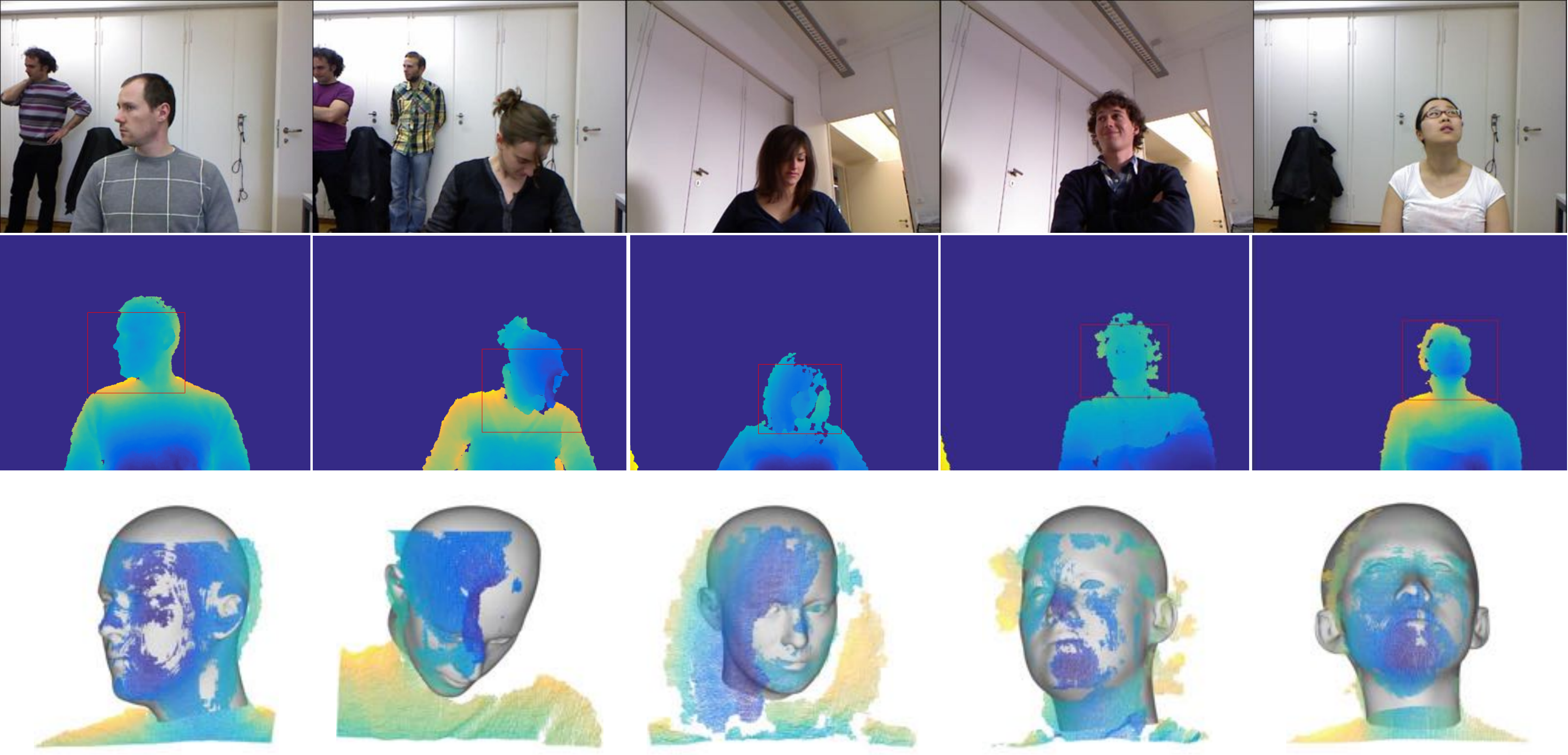}}
~
\subfloat[ICT-3DHP dataset]{\includegraphics[height=0.30\linewidth]{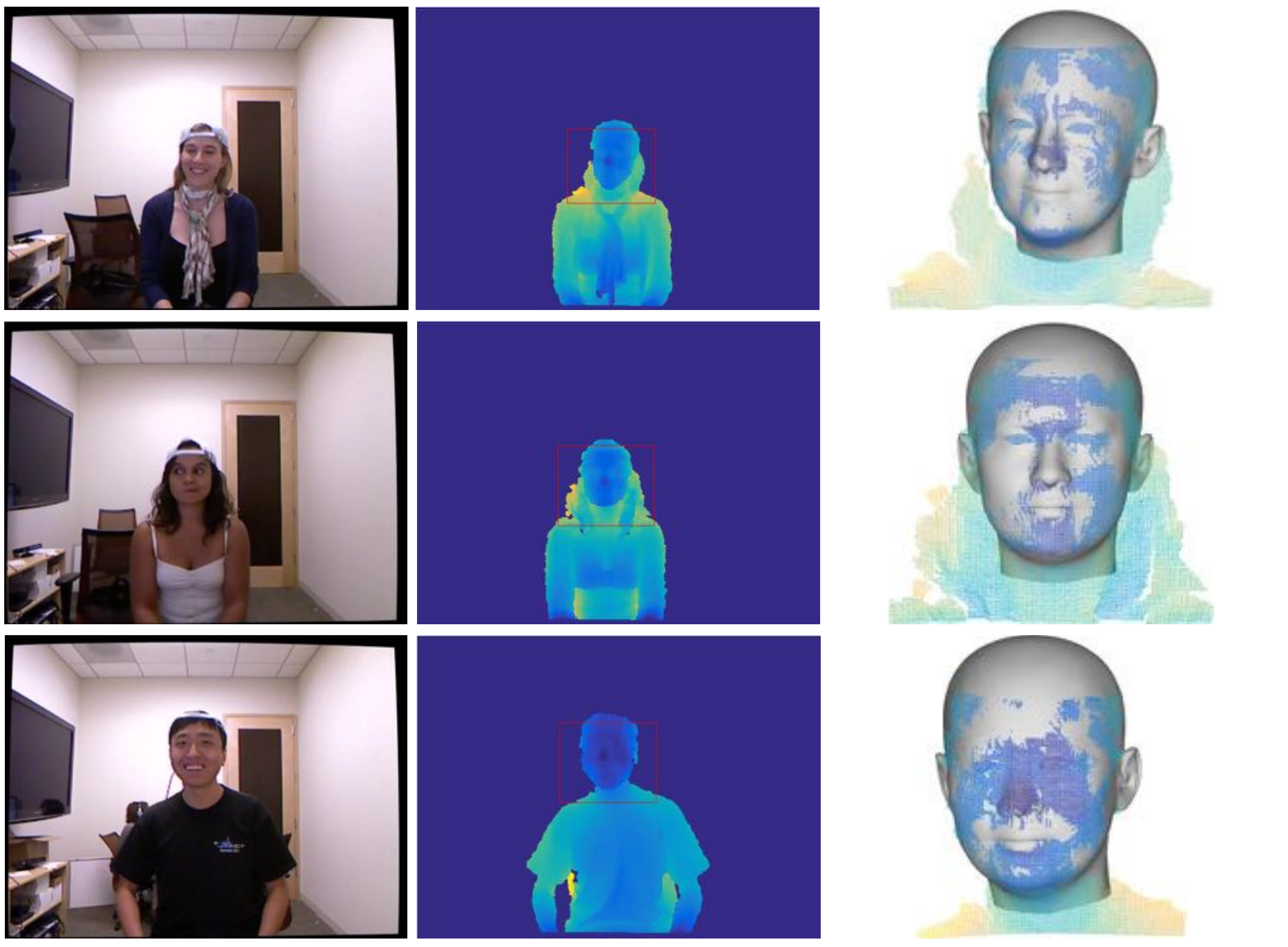}}
\caption{(a) Tracking results on the (a) Biwi and (b) ICT-3DHP dataset with the personalized face models. Our system is robust to the profiled faces and occlusions, and is also effective to facial expression variations.}
\label{fig:facial_pose_examples}
\end{figure*}

\begin{figure}
\centering
\includegraphics[width=0.9\linewidth]{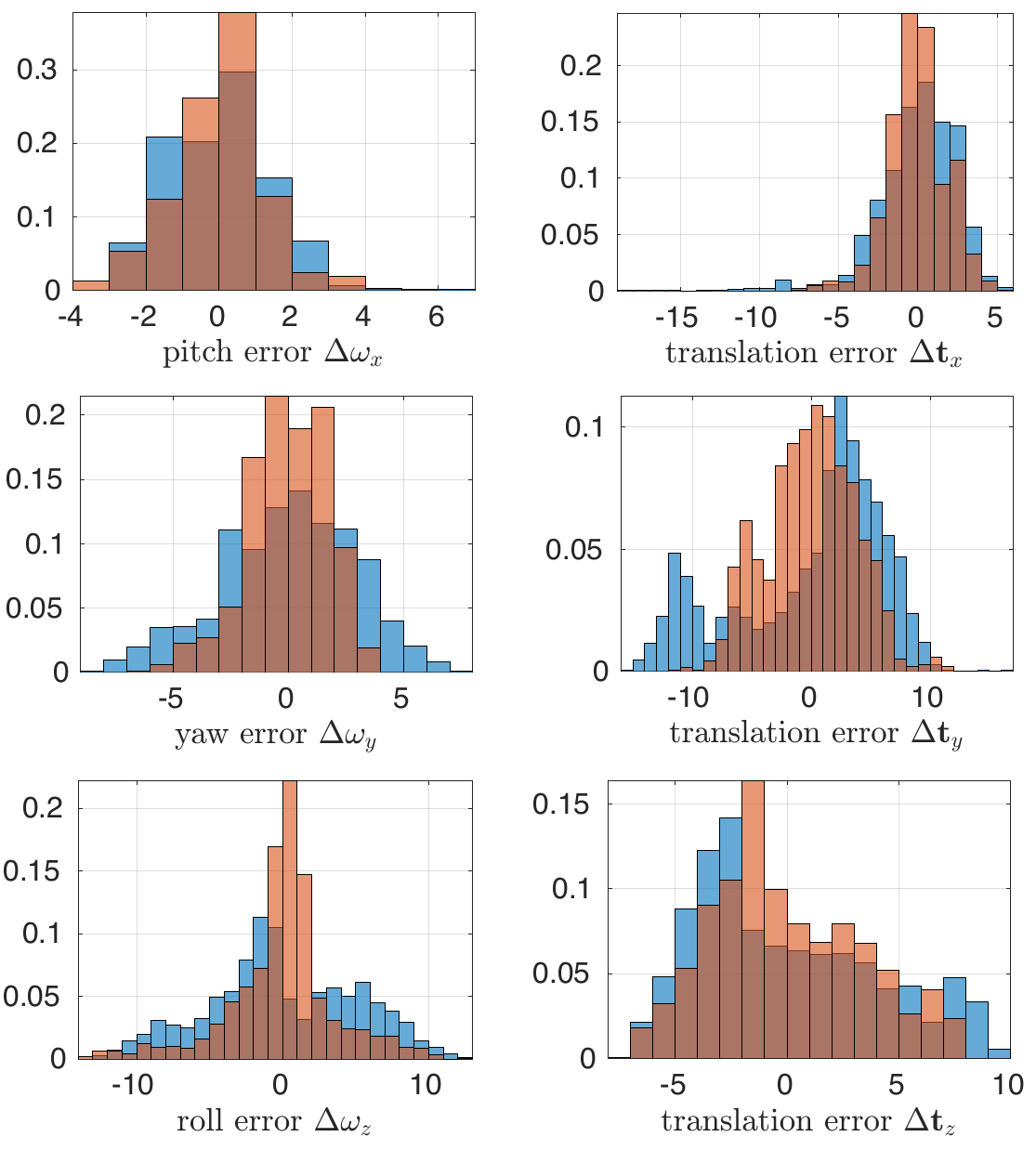}
\caption{The pose accuracy comparisons between generic and personalized face models, in terms of statistics of rotation and translation errors. The left column shows the histograms of angle errors, and the right column visualizes the histograms of translation errors. The brown bars indicate the performance by the personalized face model while the blue ones are for the generic face model.}
\label{fig:quantitative_personalized_versus_generic_face_models}
\end{figure}

\begin{figure}
\centering
\includegraphics[width=\linewidth]{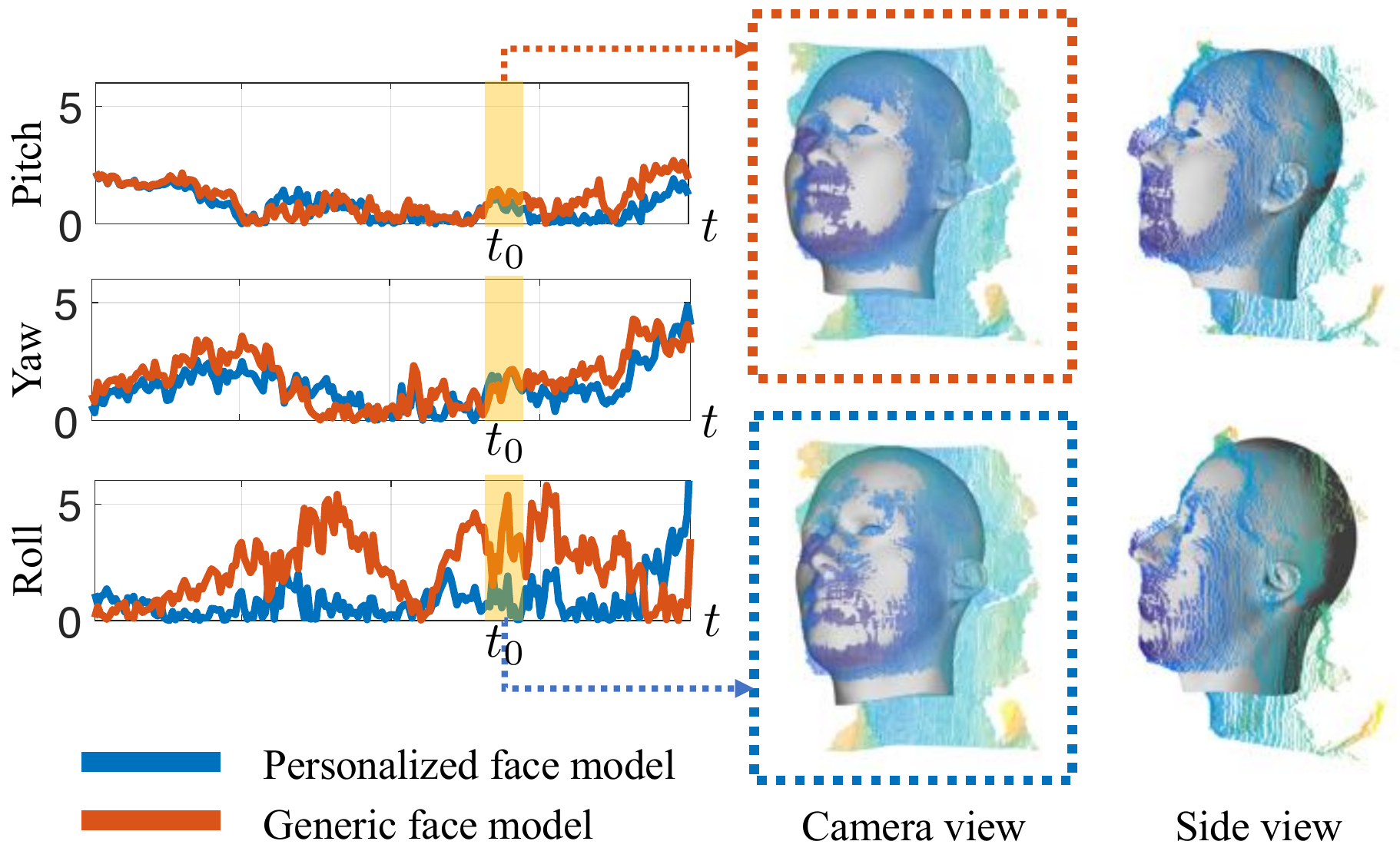}
\caption{{Profiling one short depth video clip, the absolute angle errors (with respect to pitch, yaw and roll) along the time axis, obtained by generic or personalized face model, are visualized in the left panel.
The curves by the generic face model tend to produce more errors, especially for the roll angles. By picking one frame at $t_0$, we show the face fitting results in the camera view for both models. 
When rotating to the side view, the personalized face model compactly fits the point cloud, while the generic face model fails to fit some essential regions such as the nose. The tester is also shown in Fig.~\ref{fig:exemplar_online_identity_adaptation}.}}
\label{fig:qualitative_personalized_versus_generic_face_models}
\end{figure}

\subsection{Ablation Study}
\label{sub:ablation_study}

\subsubsection{Facial Poses with Generic Face Models}
\label{ssub:facial_poses_estimation_with_generic_face_model}

The ray visibility constraint driven rigid facial pose is robust to noise, occlusions and moderate face deformations.
For example, the point clouds in Fig.~\ref{fig:rigid_pose_estimation_examples} are noisy with missing measurements and quantization errors, as well as heavy occlusions (hairs in (a) and fingers in (c)) and severe partial face scans ((a) and (b)), but the proposed method is still able to render successful poses to fit the point clouds even based on the generic face model and na\"ive pose initialization.

{The proposed facial pose estimation outperforms previous methods such as iterative closest points (ICP)~\cite{rusinkiewicz2001efficient}, given coarse initial poses proposed by Meyer's face localization method~\cite{meyer2015robust}.}
{ICP iteratively revises the transformation between the face model and the point clouds according to point-to-plane distance between matched points.}
As shown in Fig.~\ref{fig:rigid_pose_estimation_example_comparison}, the proposed method only needs the set of rays $\mathcal{V} = \{ \vec{v}_{\mathbf{q}_n} \}_{n=1}^{N_\mathcal{M}}$ but does not require explicit correspondences during estimation.
In contrast, ICP and its variants are not able to check the visibility of each matched point pair, and thus cannot guarantee a reasonable pose.
For example, as shown in Fig.~\ref{fig:rigid_pose_estimation_example_comparison}(d), ICP matches the face model with the hairs but has not been aware of the fact that the face cannot occlude the point cloud.
Moreover, the ray visibility score (RVS) is less vulnerable to bad local minima, since it rewards a complete overlap of the surface distributions $p_\mathcal{P}(\mathbf{y})$ and $p_{\mathcal{Q}\rightarrow\mathcal{P}}(\mathbf{y};\boldsymbol\theta)$, which is much less sensitive than some point estimate methods like maximum likelihood (ML) or maximum a posteriori (MAP).
For example, maximizing the likelihood $p_{\mathcal{Q}\rightarrow\mathcal{P}}(\mathbf{y};\boldsymbol\theta)$ in Eq.~\eqref{eq:signed_distance_distribution} may just seek a local mode that fails to catch the major mass of the distribution, as shown in Fig.~\ref{fig:rigid_pose_estimation_example_comparison}(e).
On the contrary, the Kullback-Leibler divergence in RVS ensures the face model distribution with the optimal $\boldsymbol\theta$ covers the bulk of information conveyed in $p_\mathcal{P}(\mathbf{y})$.
Fig.~\ref{fig:rigid_pose_estimation_examples} and~\ref{fig:rigid_pose_estimation_example_comparison} both reveal the superiority of the RVS and RVS+PSO methods in handling unconstrained facial poses with large rotations and and heavy occlusions, even with generic face model, in which the particle swarm optimization refines the facial poses one step further.

\begin{figure*}
\centering
\includegraphics[width=0.48\linewidth]{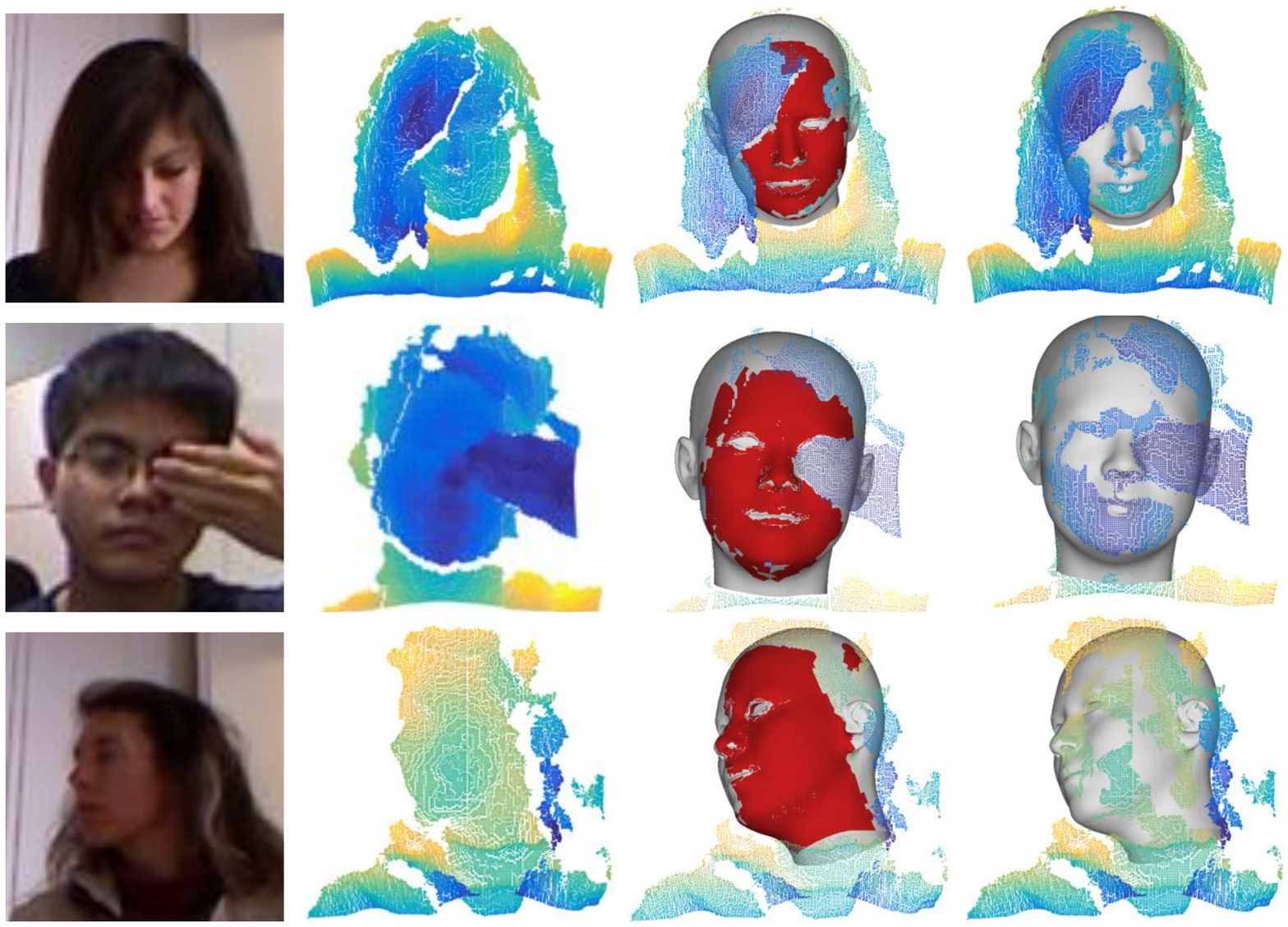}
\includegraphics[width=0.48\linewidth]{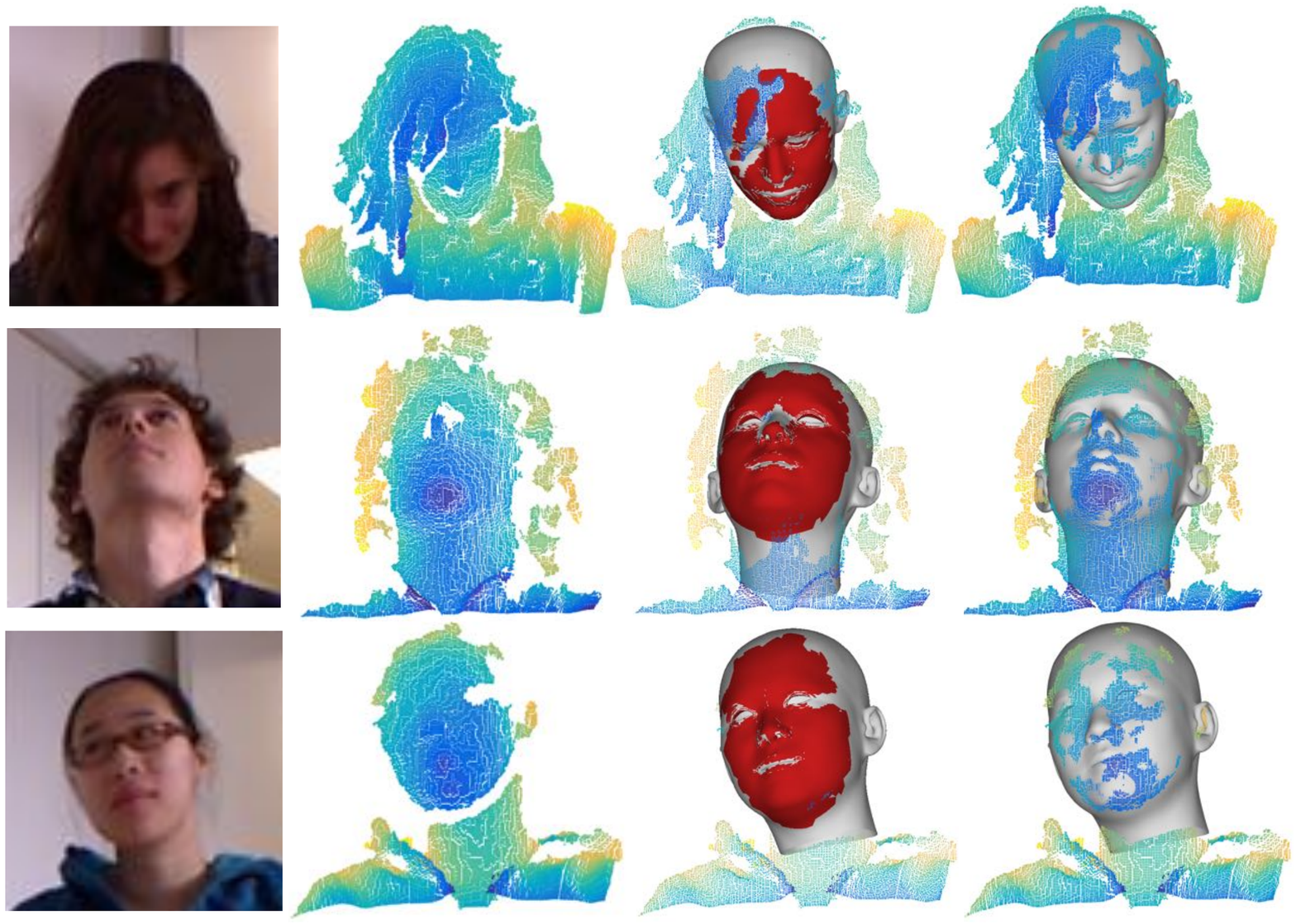}
\caption{Examples of facial pose results with the visibility detection. The third column shows the visibility masks. The last column shows the personalized face models overlaid on the point clouds.}
\label{fig:poses_with_visibility_check}
\end{figure*}

\subsubsection{Facial Poses with Personalized Face Models}
\label{ssub:facial_poses_with_personalized_face_models}

Fig.~\ref{fig:facial_pose_examples} shows some tracking results on Biwi and ICT-3DHP datasets based on the gradually adapted face models.
Although using generic model can already achieve good performance over challenging cases, as shown in Fig.~\ref{fig:rigid_pose_estimation_examples} and~\ref{fig:rigid_pose_estimation_example_comparison}, using personalized face model receives even better results both in the rotation and translation metrics.
As visualized in Fig.~\ref{fig:quantitative_personalized_versus_generic_face_models}, by comparing the angle and translation error histograms, it shows that both the angle and the translation errors by the personalized model are smaller than those by the generic model, since the personalized model tends to produce much narrower histograms with much smaller outliers from large errors, suggesting that the personalized face model indeed benefits the face tracking.

Moreover, the personalized face shape distribution enables the face model to fit compactly with the input point cloud so as to better capture some challenging poses than the generic face model (in Fig.~\ref{fig:facial_pose_examples}(a)), while the personalized expression distribution makes the estimated facial pose robust to changes in the expression space (in Fig.~\ref{fig:facial_pose_examples}(b)).
Angle error curves in Fig.~\ref{fig:qualitative_personalized_versus_generic_face_models} demonstrate the superiority of the personalized face model in eliminating more angle errors that relate to the shape mismatches between the face model and the point cloud.
For the probe frame $t_0$, both the generic and personalized face models have small pitch and yaw errors.
Despite that both models seem to produce good fitting results in the camera view, the mismatching between the generic face model and the point clouds leads to a much larger roll error and make the model fail to fit some essential facial regions such as the nose tip, as visualized in the side view.
In contrast, the personalized face model is less vulnerable to the profiled faces.

\begin{figure}[t]
\centering
\includegraphics[width=0.9\linewidth]{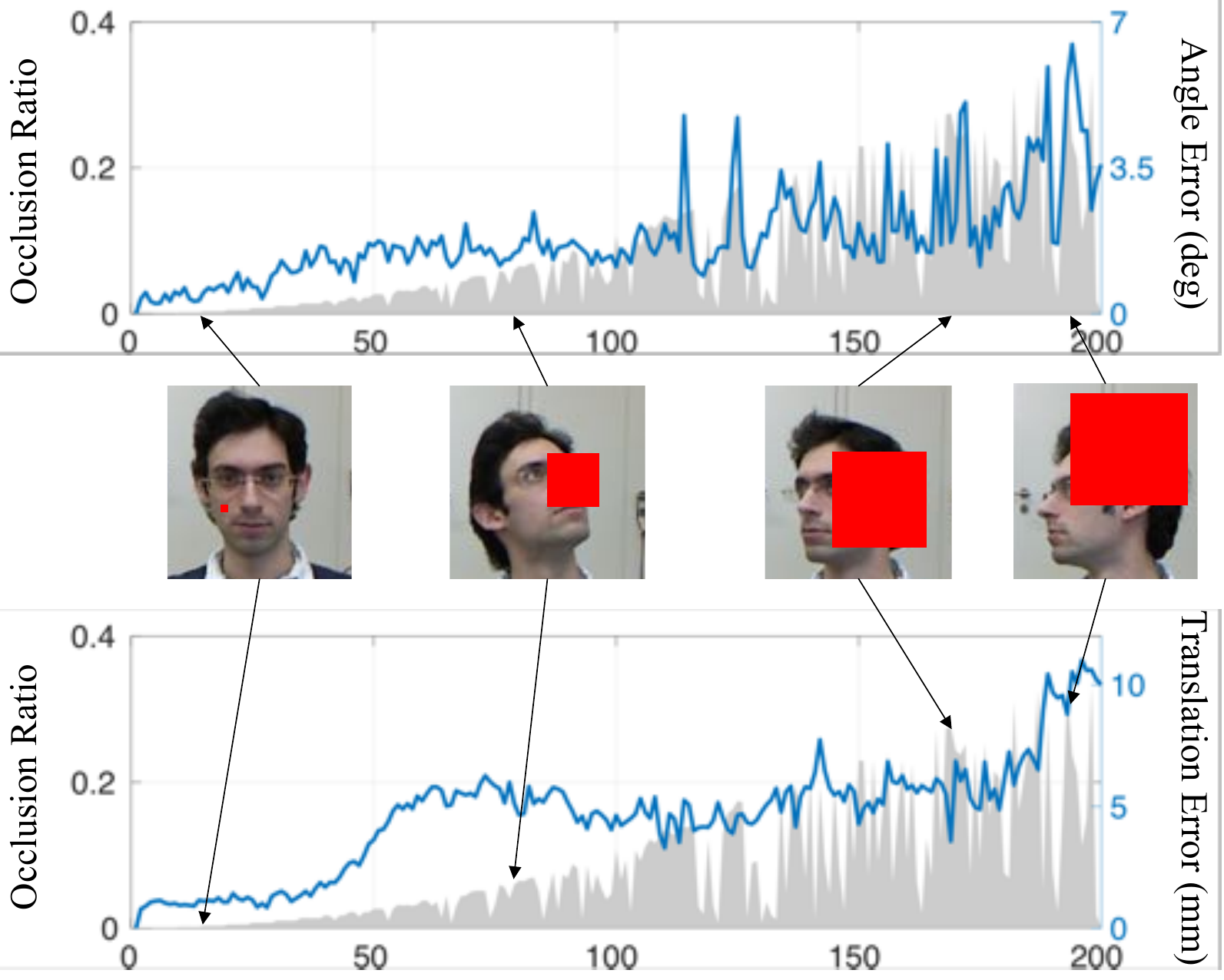}
\caption{Our facial pose tracking in handling different levels of occlusions. The gray area indicates the ratio of occlusions. The blue curves are angle errors in the upper figure and the translation errors in the lower figure. The occlusions are visualized as red masks placed in the color images (note that the occlusions are actually placed in the depth images).}
\label{fig:occlusion_evaluations}
\end{figure}

\begin{figure}[t]
\centering
\includegraphics[width=\linewidth]{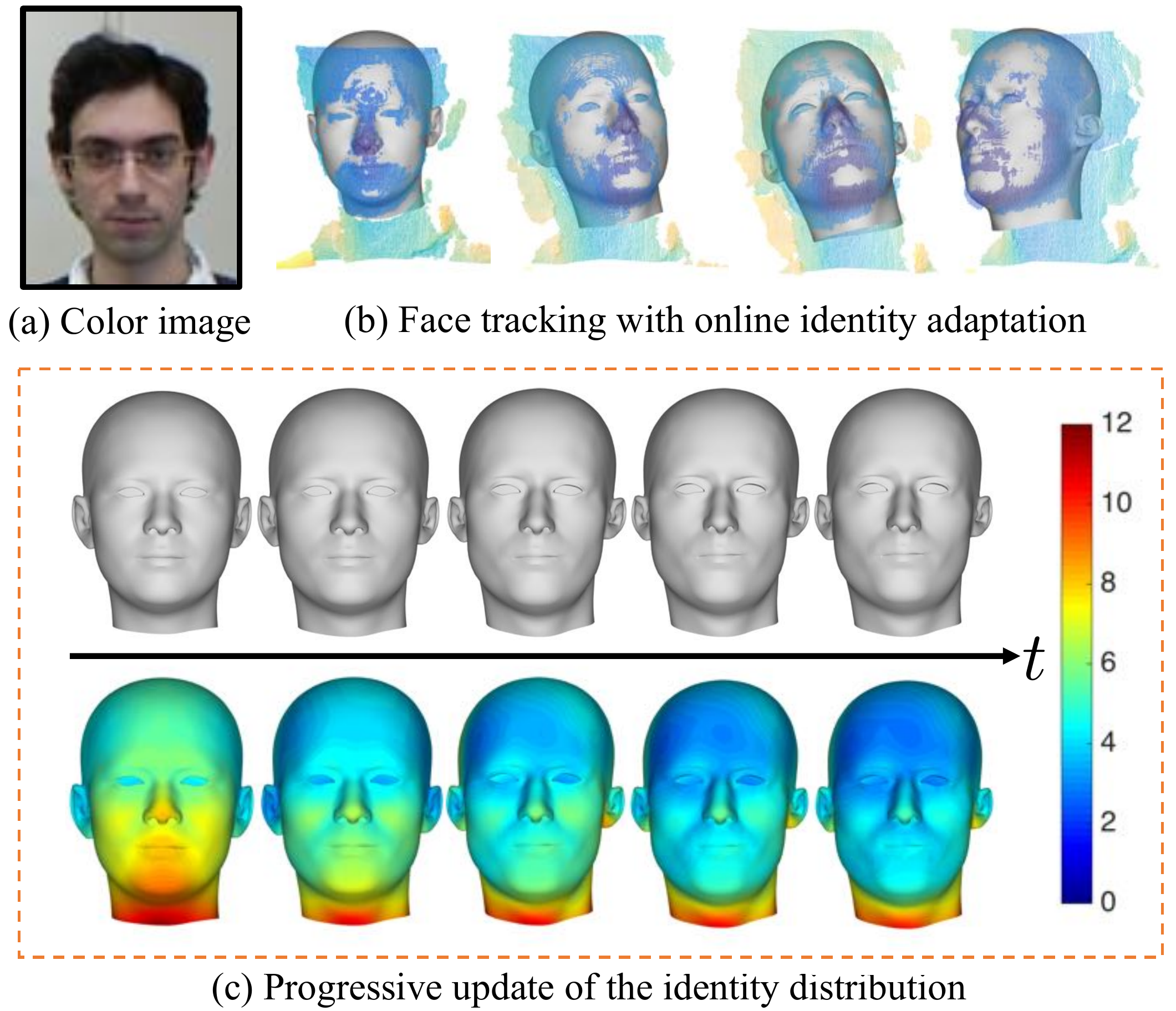}
\caption{Online identity adaptation for one exemplar Biwi tester. (a) Color image of the tester. (b) Rigid face tracking with online identity adaptation. (c) The progressive update of the parameters of the identity distribution. The upper row is the evolution of the mean shape encoded by $\boldsymbol\mu_\text{id}$ and the lower row is the evolution of the $\boldsymbol\Sigma_\text{id}$, which is visualized by the point-wise standard deviation from $\boldsymbol\Sigma_\mathcal{I}$. The variance of identity is gradually diminished with increased number of samples.}
\label{fig:exemplar_online_identity_adaptation}
\end{figure}

\begin{figure}[t]
\centering
\includegraphics[width=\linewidth]{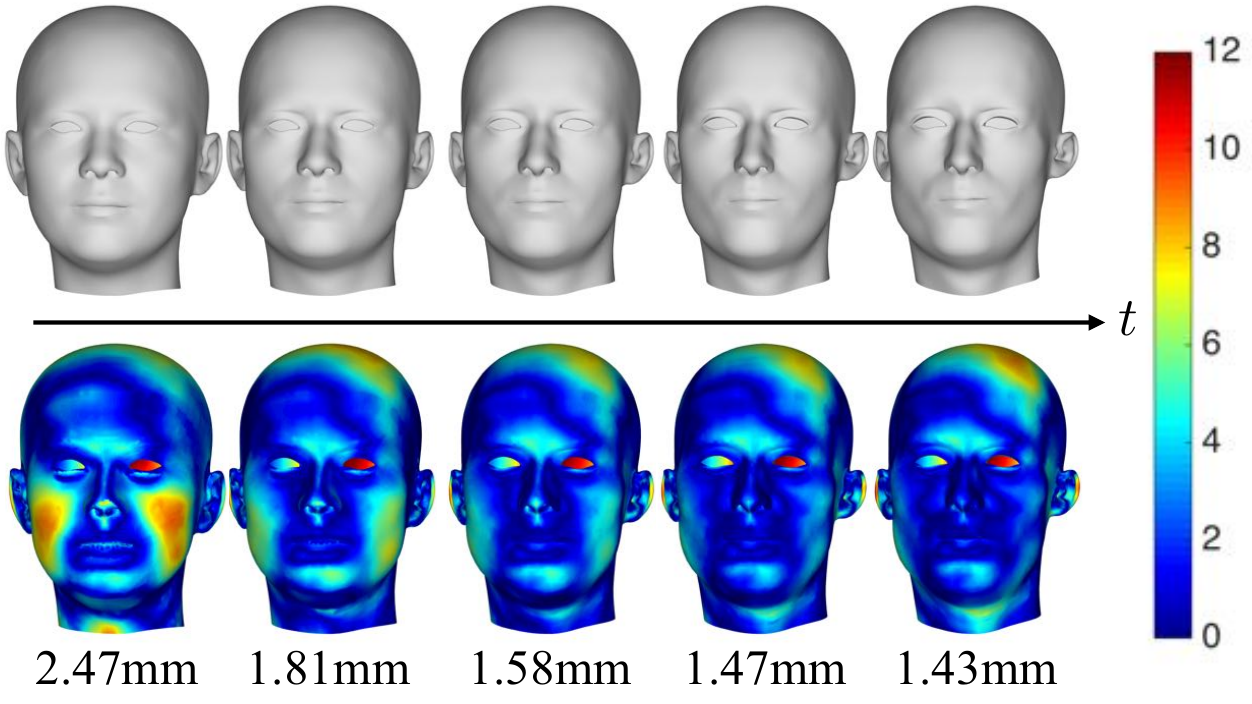}
\caption{Our identity adaptation is gradually reconstructing the tester's facial geometry. The tester is the same as that in Fig.~\ref{fig:exemplar_online_identity_adaptation}. The upper row shows the adapted face meshes, and the lower row shows per-vertex reconstruction errors and the overall \emph{facial} reconstruction errors.}
\label{fig:identity_reconstruction}
\end{figure}

\begin{figure}[t]
\centering
\includegraphics[width=\linewidth]{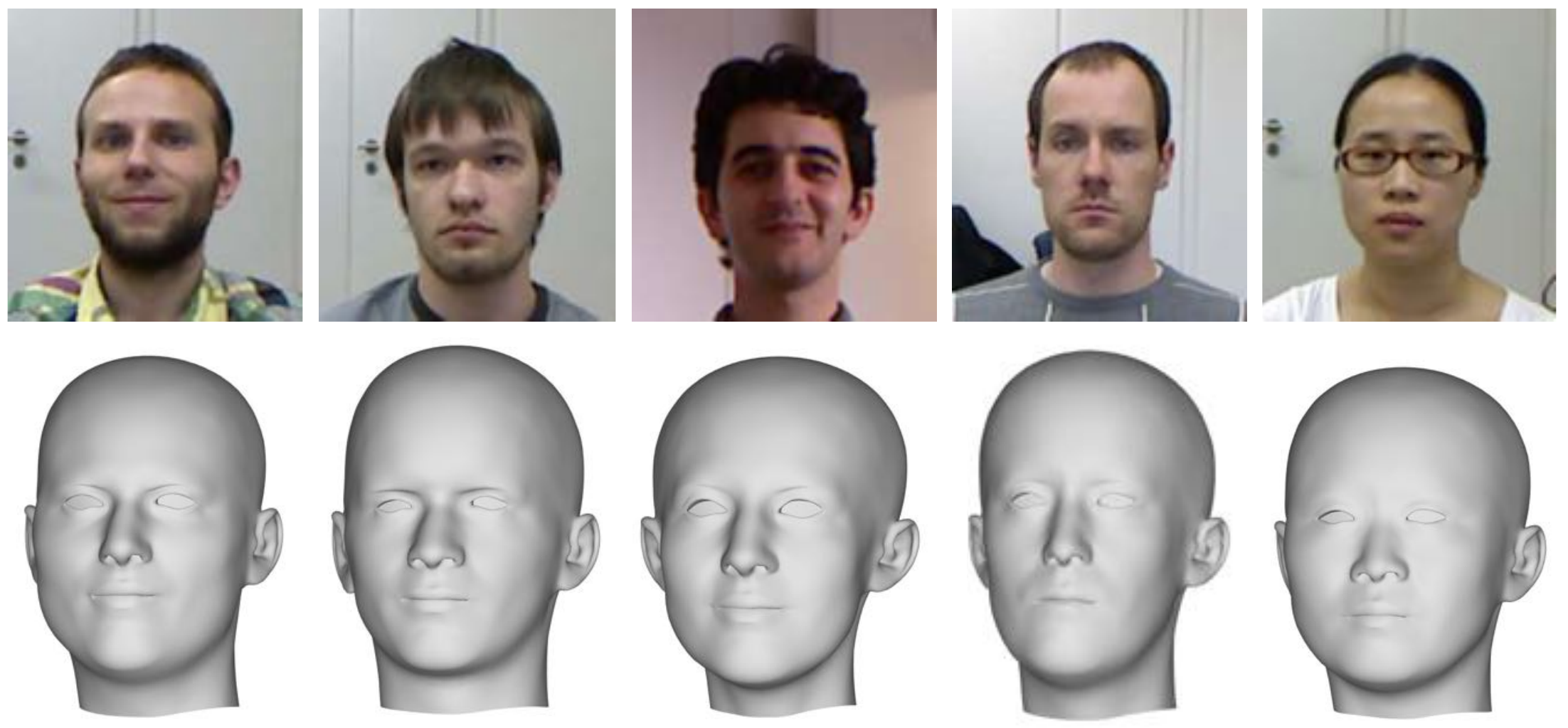}
\caption{Examples of identity adaptation. Our method successfully adapts the generic model to different identities.}
\label{fig:face_model_examples}
\end{figure}

\begin{figure*}[t]
\includegraphics[height=0.27\linewidth]{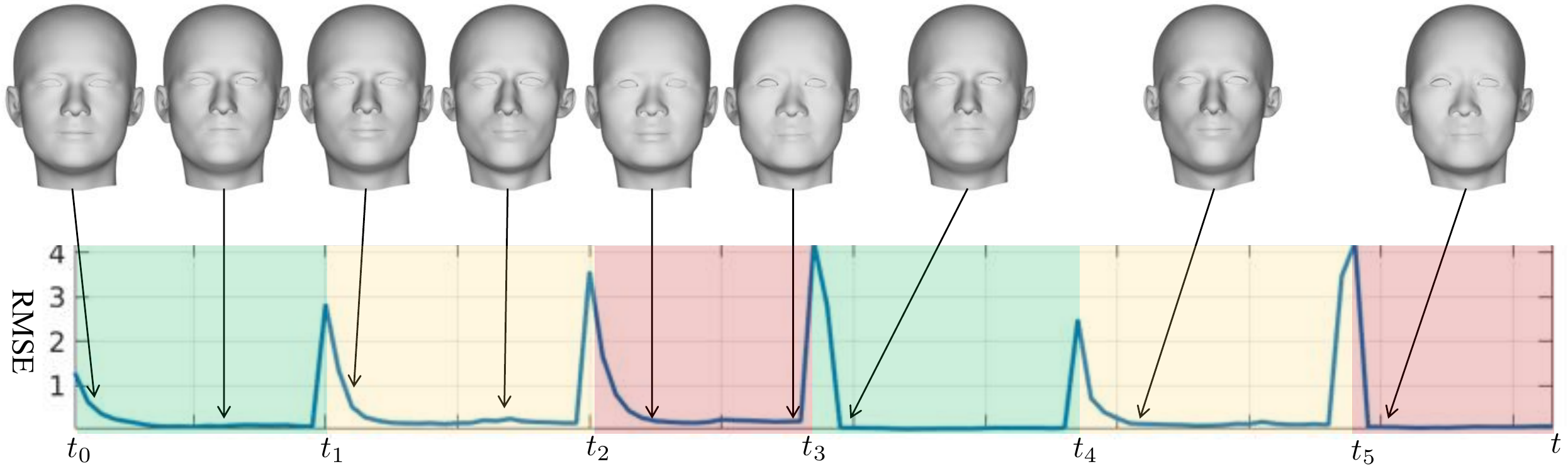}~\includegraphics[height=0.27\linewidth]{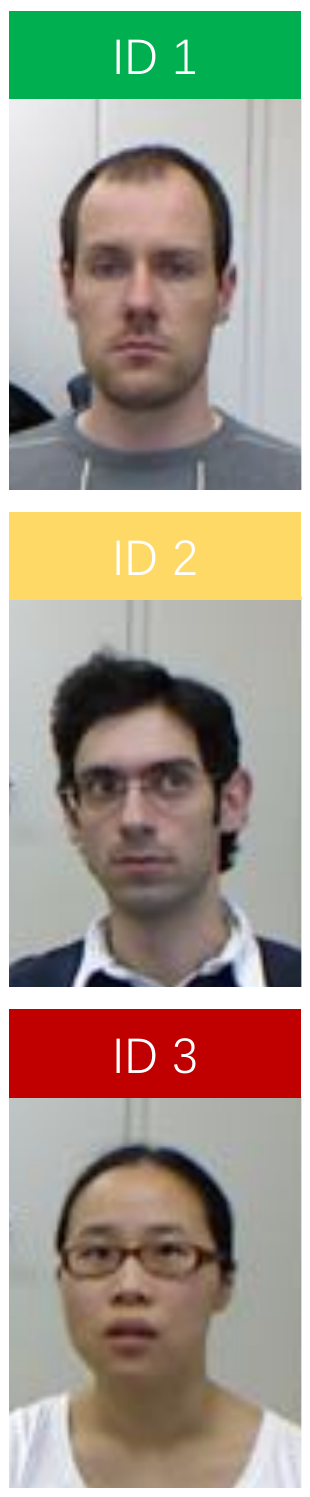}
\caption{
The proposed system can automatically switch the presented face model from one identity to another, while staying updating one identity model if its associated depth frames keep inputting into the system.
The curve shows the root mean square errors (RMSE) between the presented face models in two successive frames.
The test video concatenates six depth video clips of three identities in order, in which the green section indicates ID~1, the yellow section refers to ID~2 and the red one means ID~3.
The presented face model smoothly updates from the generic model to different identities as the RMSE curve segments in the first three sections are gradually degraded to converge, and the presented face models are gradually personalized.
The error pulses between two sections indicate the success of creating novel identity model.
Moreover, the last three sections represent the success of the fast identity switching from the stored personalized identity models.}
\label{fig:switch_user}
\end{figure*}

\subsubsection{Visibility Detection and Occlusion Handling}
\label{ssub:facial_poses_with_visibility_detection}

Meanwhile, our method efficiently infers the occlusions, partially owing this success to the reliable visibility detection embedded in the ray visibility constraint.
As shown in Fig.~\ref{fig:poses_with_visibility_check} and Fig.~\ref{fig:introduction}(a), the proposed method has checked the visibility of a face model with respect to the input point cloud such that the visibility mask tightly covers the visible regions of a face model and rejects various occlusions placed in its front.
Therefore, the proposed pose estimation takes the reliable facial points into consideration and is thus robust to severe occlusions, \eg, self-occlusions like profiled faces, accessories, hands and \etc.

We also challenge our robust facial pose tracking with different levels of occlusions, similarly as the way used in Hsieh~\etal~\cite{hsieh2015unconstrained}.
The synthesized occlusion regions have gradually increased sizes and they are randomly placed around the face center.
The tracking accuracy is measured on the Biwi dataset, as visualized in Fig.~\ref{fig:occlusion_evaluations}.
The proposed visibility constraint can handle a large amount of occlusions, and the tracking errors will not increase tremendously.

\subsubsection{Online Face Model Adaptation}
\label{ssub:online_face_model_adaptation}

The proposed method provides an online identity adaptation that progressively adapts the face model to one test subject as shown in Fig.~\ref{fig:exemplar_online_identity_adaptation}.
More personalized face models are also visualized in Fig.~\ref{fig:face_model_examples}.
The statistical face model has the ability to cover various identities ranging from different ages, races and genders.
We also compare the refined face models to the groundtruth face meshes in the Biwi datasets, as visualized in Fig.~\ref{fig:identity_reconstruction}.
As more frames coming into the system, the personalized face models will gradually have smaller shape differences (termed as point-to-plane cost for matched 3D points) to the groundtruth meshes. 
Note that the speed of convergence depends on the ratio of occlusions with respect to the facial region, and fewer occlusions lead to faster adaptation.
{
But the adaptation quality is sometimes contaminated due to falsely aggregating the occluded face regions (short hair, beard and so on) into our model updating system, thus the personalized face models may suffer from distortions around the side heads, as shown in Fig.~\ref{fig:face_model_examples}.
}

The proposed online switch scheme is able to either instantaneously create a novel identity model for an unseen tester, or switch different stored personalized models according to the change of active testers.
As visualized in Fig.~\ref{fig:switch_user}, six depth video clips capturing three different identities (marked by ID~$1$ to $3$) from Biwi dataset are put into the proposed system.
Initially, the first identity model is generated by the generic model, and it is gradually warped to the shape of ID~$1$ given a series of depth frames referring to ID~$1$.
Since the subsequent frames after $t_1$ capture a novel identity ID~$2$, the proposed system automatically adds a novel identity model according to the MAP estimation from Eq.~\eqref{eq:switch_posterior}.
The presented face model is replaced by that of ID~$2$, and it is updated based on the inputs frames about ID~$2$.
Same process also happens to ID~$3$ when its frames are put into the system after $t_2$.
Interestingly, the proposed system can quickly parse a suitable identity model for the input depth frame from the stored personalized identity models, based on a similar switching criterion from Eq.~\eqref{eq:switch_posterior}.
For example, ID~$1$ is immediately parsed at time $t_3$, and the presented face model does not receive meaningful updates a step forward since this identity has already been well personalized and the additional frames have not introduced novel knowledges about its facial shape.
ID~$2$ and ID $3$ are also effectively parsed when the right identities are captured again, but we notice that ID~$2$ still receives incremental shape adaptation as additional informative frames are helpful to adapt its identity model.
Therefore, the proposed online identity adaptation offers a promising ability to instantaneously switch from one subject to a new one followed by a smooth facial identity model updating.

\subsubsection{Execution Efficiency}
\label{ssub:execution_efficiency}

The proposed RVS-based facial pose estimation has a comparable complexity as ICP (in Tab.~\ref{tab:componentwise_runtime_comparison}).
Thanks to the analytical KL-divergence for Gaussian distributions, the ray visibility score is analytical and contains a similar point-wise squared cost term as ICP.
The added temporal coherence term requires a smaller computational budget than RVS and the scale accumulation only adds a marginal costs in computation ((RVS+TE in Tab.~\ref{tab:componentwise_runtime_comparison}).
PSO is indeed the bottleneck of our system (see RVS+TE+PSO in Tab.~\ref{tab:componentwise_runtime_comparison}), but as it is optionally added to tackle tracking failure, it will not tremendously slow down the executive speed.
Since the identity adaptation requires a relative large linear solver in estimating $\mathbf{w}_\text{id}$, it (IA in Tab.~\ref{tab:componentwise_runtime_comparison}) costs a scale more computational budgets than the tracking module (RVS+TE).

\begin{table}[h]
\vspace{-3mm}
\centering
\caption{Component-wise Runtime Comparison (MATLAB Platform)}
\label{tab:componentwise_runtime_comparison}
\vspace{-3mm}
\begin{tabular}{c|c||ccc|c}
\hline
 &  ICP & RVS &  RVS+TE & RVS+TE+PSO & IA \\
\hline
$\Delta t$ (sec) & 0.0156 & 0.0193 & 0.0233 & 0.328 & 0.131 \\
\hline
\end{tabular}
\vspace{-3mm}
\end{table}

\subsection{Quantitative Comparisons with the prior arts}

\begin{table}[b]
\centering
\caption{Evaluations on Biwi dataset}
\label{tab:evaluation_biwi}
\vspace{-3mm}
\begin{tabular}{c|c|c|c|c}
\hline
\multirow{2}{*}{Method} & \multicolumn{4}{c}{Errors} \\
\cline{2-5}
& Yaw ($^\circ$) & Pitch ($^\circ$) & Roll ($^\circ$) & Trans (mm) \\
\hline
Ours & \textbf{1.6} & \textbf{1.7} & \textbf{1.8} & \textbf{5.5} \\
\hline
Sheng~\etal~\cite{Sheng_2017_CVPR} & 2.3 & 2.0 & 1.9 & 6.9 \\
RF~\cite{fanelli2011real} & 8.9 & 8.5 & 7.9 & 14.0 \\
Martin~\cite{martin2014real} & 3.6 & 2.5 & 2.6 & 5.8 \\
CLM-Z~\cite{baltruvsaitis20123d} & 14.8 & 12.0& 23.3& 16.7 \\
TSP~\cite{papazov2015real} & 3.9 & 3.0 & 2.5 & 8.4 \\
PSO~\cite{padeleris2012head} & 11.1 & 6.6 & 6.7 & 13.8 \\
Meyer~\etal~\cite{meyer2015robust} &  2.1 &  2.1 &  2.4 & 5.9 \\
Li~\etal$^\star$~\cite{li2015real} & 2.2 & 1.7 & 3.2 & $-$\\
\hline
\end{tabular}
\end{table}

We compare our final model with a number of prior methods~\cite{fanelli2011real,meyer2015robust,martin2014real,baltruvsaitis20123d,papazov2015real,padeleris2012head,li2015real,Sheng_2017_CVPR} for depth-based 3D facial pose tracking on the Biwi~\cite{fanelli2011real} and ICT-3DHP~\cite{baltruvsaitis20123d} datasets.
Tab.~\ref{tab:evaluation_biwi} shows the average absolute errors for the rotation angles and the average Euclidean errors for the translation on the Biwi dataset. 
The rotational errors were further quantified with respect to the yaw, pitch and roll angles, respectively. 
Similarly in Tab.~\ref{tab:evaluation_ict_3dhp}, we evaluate the average rotation errors on the ICT-3DHP dataset, as translations are unavailable for ICT-3DHP datasets~\cite{baltruvsaitis20123d}.
Note that the results of the reference methods are taken directly from those reported by their respective authors in literature.

\begin{table}[t]
\centering
\caption{Evaluations on ICT-3DHP dataset}
\label{tab:evaluation_ict_3dhp}
\vspace{-3mm}
\begin{tabular}{c|c|c|c}
\hline
\hline
\multirow{2}{*}{Method} & \multicolumn{3}{c}{Errors} \\
\cline{2-4}
& Yaw ($^\circ$) & Pitch ($^\circ$) & Roll ($^\circ$)\\
\hline
Ours & \textbf{2.5} & \textbf{3.0} & \textbf{2.7} \\
\hline
Sheng~\etal~\cite{Sheng_2017_CVPR} & 3.4 & 3.2 & 3.3 \\
RF~\cite{fanelli2011real} & 7.2 & 9.4 & 7.5 \\
CLM-Z~\cite{baltruvsaitis20123d} & 6.9 & 7.1 & 10.5\\
Li~\etal$^\star$~\cite{li2015real} & 3.3 & 3.1 & 2.9\\
\hline
\hline
\end{tabular}
\end{table}

On the Biwi dataset, the proposed method produces the overall lowest rotation errors among the depth-based head pose tracking methods~\cite{Sheng_2017_CVPR,fanelli2011real,baltruvsaitis20123d,martin2014real,papazov2015real,li2015real,padeleris2012head,meyer2015robust}. 
Although no appearance information is used, the proposed approach outperforms the existing state-of-the-art method~\cite{li2015real} (marked with $\star$ in Tab.~\ref{tab:evaluation_biwi} and~\ref{tab:evaluation_ict_3dhp}) that employed both RGB and depth data.
Similar conclusions can also be drawn on the ICT-3DHP dataset, where the proposed method also achieves a superior performance on estimating the rotation parameters in comparison with the random forests~\cite{fanelli2011real} and CLM-Z~\cite{baltruvsaitis20123d}. 
Our performance is slightly superior to Li~\cite{li2015real} even without using any color information.

In comparison with the previously released conference paper~\cite{Sheng_2017_CVPR}, the superiority of our proposed method attributes to more effective constrains such as the temporal coherence and scale stabilization.
Therefore not only the flickering from the estimated poses among adjacent frames is eliminated, but also the tracking robustness is improved, since some ambiguous cases such as heavily occluded or profiled faces can obtain more confidences from the per-pixel facial flow across adjacent frames.

As for the translation parameters, the proposed method also achieves very competitive performance on the Biwi dataset. 
The proposed method outperforms the prior arts, especially those proposed by Meyer~\etal~\cite{meyer2015robust} and Sheng~\etal~\cite{Sheng_2017_CVPR}.
In comparison to Sheng~\etal~\cite{Sheng_2017_CVPR}, the introduction of temporal coherence and scale accumulation remarkably increases the translation accuracy, and makes the proposed method outperforms the previous state-of-the-art methods, such as Meyer~\etal~\cite{meyer2015robust}.

\subsection{Limitations}
\label{ssec:limitations}

The proposed system is inevitably vulnerable when the input depth video is contaminated by heavy noise, outliers and quantization errors.
Since effective clues like facial landmarks are inaccessible due to the missing of the color information, difficult facial poses with extreme large rotational angles or occlusions are sometimes hard to be well tracked.
Even though this problem can be relieved by enforcing temporal coherence, the results still receive accumulative tracking errors from the previous frames.
Moreover, tightly fitting occlusions, such as face masks, veils and short hair， cannot be well handled by the proposed method.

\section{Conclusions And Future Works}

We propose a robust 3D facial pose tracking for commodity depth sensors that brings about the state-of-the-art performances on popular facial pose datasets.
The proposed generative face model and the ray visibility constraint ensure a robust groupwise 3D facial pose tracking accounting for all inherent identities and expressions that effectively handles heavy occlusions, profiled faces and expression variations.
The online switchable identity adaptation is able to gradually personalized the face model for one specific user, and instantaneously switch among stored identities and create a new identity model for a novel user.

Some future works are beneficial:
effective long-term temporal coherence still deserves attention since it provides smoother and more complex tracking trajectories.
Moreover, effective depth-based features are helpful to provide semantic correspondences to eliminate trivial or outlier solutions.

\ifCLASSOPTIONcompsoc
  \section*{Acknowledgments}
\else
  \section*{Acknowledgment}
\fi

This research is supported by the BeingTogether Centre, a collaboration between Nanyang Technological University (NTU) Singapore and University of North Carolina (UNC) at Chapel Hill. The BeingTogether Centre is supported by the National Research Foundation, Prime Minister’s Office, Singapore under its International Research Centres in Singapore Funding Initiative. This work is also in part supported MoE Tier-2 Grant (2016-T2-2-065) of Singapore.

\ifCLASSOPTIONcaptionsoff
  \newpage
\fi



\bibliographystyle{IEEEtran}
\bibliography{IEEEabrv,lshengrefs}

\begin{thebibliography}{10}
\providecommand{\url}[1]{#1}
\csname url@samestyle\endcsname
\providecommand{\newblock}{\relax}
\providecommand{\bibinfo}[2]{#2}
\providecommand{\BIBentrySTDinterwordspacing}{\spaceskip=0pt\relax}
\providecommand{\BIBentryALTinterwordstretchfactor}{4}
\providecommand{\BIBentryALTinterwordspacing}{\spaceskip=\fontdimen2\font plus
\BIBentryALTinterwordstretchfactor\fontdimen3\font minus
  \fontdimen4\font\relax}
\providecommand{\BIBforeignlanguage}[2]{{%
\expandafter\ifx\csname l@#1\endcsname\relax
\typeout{** WARNING: IEEEtran.bst: No hyphenation pattern has been}%
\typeout{** loaded for the language `#1'. Using the pattern for}%
\typeout{** the default language instead.}%
\else
\language=\csname l@#1\endcsname
\fi
#2}}
\providecommand{\BIBdecl}{\relax}
\BIBdecl

\bibitem{cao20133d}
C.~Cao, Y.~Weng, S.~Lin, and K.~Zhou, ``3d shape regression for real-time
  facial animation,'' \emph{ACM Trans. Graph.}, vol.~32, no.~4, p.~41, 2013.

\bibitem{Egger2018}
B.~Egger, S.~Sch{\"o}nborn, A.~Schneider, A.~Kortylewski, A.~Morel-Forster,
  C.~Blumer, and T.~Vetter, ``Occlusion-aware 3d morphable models and an
  illumination prior for face image analysis,'' \emph{Int. J. Comput. Vis.},
  Jan 2018.

\bibitem{Booth_2017_CVPR}
J.~Booth, E.~Antonakos, S.~Ploumpis, G.~Trigeorgis, Y.~Panagakis, and
  S.~Zafeiriou, ``3d face morphable models "in-the-wild",'' in \emph{Proc. IEEE
  Conf. Comput. Vis. Pattern Recognit.}\hskip 1em plus 0.5em minus 0.4em\relax
  IEEE, July 2017.

\bibitem{Jackson_2017_ICCV}
A.~S. Jackson, A.~Bulat, V.~Argyriou, and G.~Tzimiropoulos, ``Large pose 3d
  face reconstruction from a single image via direct volumetric cnn
  regression,'' in \emph{Proc. IEEE Int. Conf. Comput. Vis.}\hskip 1em plus
  0.5em minus 0.4em\relax IEEE, Oct 2017.

\bibitem{Richardson_2017_CVPR}
E.~Richardson, M.~Sela, R.~Or-El, and R.~Kimmel, ``Learning detailed face
  reconstruction from a single image,'' in \emph{Proc. IEEE Conf. Comput. Vis.
  Pattern Recognit.}\hskip 1em plus 0.5em minus 0.4em\relax IEEE, July 2017.

\bibitem{saito2016real}
S.~Saito, T.~Li, and H.~Li, ``Real-time facial segmentation and performance
  capture from {RGB} input,'' \emph{Proc. Euro. Conf. Comput. Vis.}, 2016.

\bibitem{Dou_2017_CVPR}
P.~Dou, S.~K. Shah, and I.~A. Kakadiaris, ``End-to-end 3d face reconstruction
  with deep neural networks,'' in \emph{Proc. IEEE Conf. Comput. Vis. Pattern
  Recognit.}\hskip 1em plus 0.5em minus 0.4em\relax IEEE, July 2017.

\bibitem{cao2014displaced}
C.~Cao, Q.~Hou, and K.~Zhou, ``Displaced dynamic expression regression for
  real-time facial tracking and animation,'' \emph{ACM Trans. Graph.}, vol.~33,
  no.~4, p.~43, 2014.

\bibitem{cao2015real}
C.~Cao, D.~Bradley, K.~Zhou, and T.~Beeler, ``Real-time high-fidelity facial
  performance capture,'' \emph{ACM Trans. Graph.}, vol.~34, no.~4, p.~46, 2015.

\bibitem{blanz1999morphable}
V.~Blanz and T.~Vetter, ``A morphable model for the synthesis of 3{D} faces,''
  in \emph{Proceedings of the 26th annual conference on Computer graphics and
  interactive techniques}.\hskip 1em plus 0.5em minus 0.4em\relax ACM
  Press/Addison-Wesley Publishing Co., 1999, pp. 187--194.

\bibitem{hsieh2015unconstrained}
P.-L. Hsieh, C.~Ma, J.~Yu, and H.~Li, ``Unconstrained realtime facial
  performance capture,'' in \emph{Proc. IEEE Conf. Comput. Vis. Pattern
  Recognit.}\hskip 1em plus 0.5em minus 0.4em\relax IEEE, 2015, pp. 1675--1683.

\bibitem{pham2016robust}
H.~X. Pham and V.~Pavlovic, ``Robust real-time 3d face tracking from rgbd
  videos under extreme pose, depth, and expression variation,'' in \emph{Proc.
  IEEE Conf. 3D Vis.}\hskip 1em plus 0.5em minus 0.4em\relax IEEE, Oct 2016,
  pp. 441--449.

\bibitem{thomas2016augmented}
D.~Thomas and R.~I. Taniguchi, ``Augmented blendshapes for real-time
  simultaneous 3d head modeling and facial motion capture,'' in \emph{Proc.
  IEEE Conf. Comput. Vis. Pattern Recognit.}\hskip 1em plus 0.5em minus
  0.4em\relax IEEE, June 2016, pp. 3299--3308.

\bibitem{weise2011realtime}
T.~Weise, S.~Bouaziz, H.~Li, and M.~Pauly, ``Realtime performance-based facial
  animation,'' in \emph{ACM Trans. Graph.}, vol.~30, no.~4.\hskip 1em plus
  0.5em minus 0.4em\relax ACM, 2011, p.~77.

\bibitem{li2013realtime}
H.~Li, J.~Yu, Y.~Ye, and C.~Bregler, ``Realtime facial animation with
  on-the-fly correctives.'' \emph{ACM Trans. Graph.}, vol.~32, no.~4, pp.
  42--1, 2013.

\bibitem{8368264}
Y.~Yu, K.~F. Mora, and J.~M. Odobez, ``Headfusion: 360 head pose tracking
  combining 3d morphable model and 3d reconstruction,'' \emph{{IEEE} Trans.
  Pattern Anal. Mach. Intell.}, pp. 1--1, 2018.

\bibitem{tan2017facecollage}
F.~Tan, C.-W. Fu, T.~Deng, J.~Cai, and T.-J. Cham, ``{FaceCollage}: A rapidly
  deployable system for real-time head reconstruction for on-the-go 3d
  telepresence,'' in \emph{Proceedings of the 2017 ACM on Multimedia
  Conference}.\hskip 1em plus 0.5em minus 0.4em\relax ACM, 2017, pp. 64--72.

\bibitem{sun2008automatic}
Y.~Sun and L.~Yin, ``Automatic pose estimation of 3d facial models,'' in
  \emph{Proc. IEEE Int. Conf. Pattern Recognit.}\hskip 1em plus 0.5em minus
  0.4em\relax IEEE, 2008, pp. 1--4.

\bibitem{breitenstein2008real}
M.~D. Breitenstein, D.~Kuettel, T.~Weise, L.~Van~Gool, and H.~Pfister,
  ``Real-time face pose estimation from single range images,'' in \emph{Proc.
  IEEE Conf. Comput. Vis. Pattern Recognit.}\hskip 1em plus 0.5em minus
  0.4em\relax IEEE, 2008, pp. 1--8.

\bibitem{papazov2015real}
C.~Papazov, T.~K. Marks, and M.~Jones, ``Real-time 3{D} head pose and facial
  landmark estimation from depth images using triangular surface patch
  features,'' in \emph{Proc. IEEE Conf. Comput. Vis. Pattern Recognit.}\hskip
  1em plus 0.5em minus 0.4em\relax IEEE, 2015, pp. 4722--4730.

\bibitem{fanelli2011real}
G.~Fanelli, T.~Weise, J.~Gall, and L.~Van~Gool, ``Real time head pose
  estimation from consumer depth cameras,'' in \emph{Pattern
  Recognition}.\hskip 1em plus 0.5em minus 0.4em\relax Springer, 2011, pp.
  101--110.

\bibitem{fanelli2011randomforests}
G.~Fanelli, J.~Gall, and L.~Van~Gool, ``Real time head pose estimation with
  random regression forests,'' in \emph{Proc. IEEE Conf. Comput. Vis. Pattern
  Recognit.}\hskip 1em plus 0.5em minus 0.4em\relax IEEE, 2011, pp. 617--624.

\bibitem{riegler2014hough}
G.~Riegler, D.~Ferstl, M.~R{\"u}ther, and H.~Bischof, ``Hough networks for head
  pose estimation and facial feature localization,'' in \emph{Proc. British
  Mach. Vis. Conf.}\hskip 1em plus 0.5em minus 0.4em\relax BMVA Press, 2014.

\bibitem{kazemi2014real}
V.~Kazemi, C.~Keskin, J.~Taylor, P.~Kohli, and S.~Izadi, ``Real-time face
  reconstruction from a single depth image,'' in \emph{Proc. IEEE Conf. 3D
  Vis.}, vol.~1.\hskip 1em plus 0.5em minus 0.4em\relax IEEE, 2014, pp.
  369--376.

\bibitem{cao2014facewarehouse}
C.~Cao, Y.~Weng, S.~Zhou, Y.~Tong, and K.~Zhou, ``Facewarehouse: A 3{D} facial
  expression database for visual computing,'' \emph{{IEEE} Trans. Vis. Comput.
  Graphics}, vol.~20, no.~3, pp. 413--425, 2014.

\bibitem{vlasic2005face}
D.~Vlasic, M.~Brand, H.~Pfister, and J.~Popovi{\'c}, ``Face transfer with
  multilinear models,'' \emph{ACM Trans. Graph.}, vol.~24, no.~3, pp. 426--433,
  2005.

\bibitem{Sheng_2017_CVPR}
L.~Sheng, J.~Cai, T.-J. Cham, V.~Pavlovic, and K.~Ngi~Ngan, ``A generative
  model for depth-based robust 3d facial pose tracking,'' in \emph{Proc. IEEE
  Conf. Comput. Vis. Pattern Recognit.}\hskip 1em plus 0.5em minus 0.4em\relax
  IEEE, July 2017.

\bibitem{Li1993motion}
H.~Li, P.~Roivainen, and R.~Forchheimer, ``3-{D} motion estimation in
  model-based facial image coding,'' \emph{{IEEE} Trans. Pattern Anal. Mach.
  Intell.}, vol.~15, no.~6, pp. 545--555, Jun 1993.

\bibitem{Black1995tracking}
M.~J. Black and Y.~Yacoob, ``Tracking and recognizing rigid and non-rigid
  facial motions using local parametric models of image motion,'' in
  \emph{Proc. IEEE Int. Conf. Comput. Vis.}\hskip 1em plus 0.5em minus
  0.4em\relax IEEE, Jun 1995, pp. 374--381.

\bibitem{DeCarlo2000opticalflow}
D.~DeCarlo and D.~Metaxas, ``Optical flow constraints on deformable models with
  applications to face tracking,'' \emph{Int. J. Comput. Vis.}, vol.~38, no.~2,
  pp. 99--127, 2000.

\bibitem{kazemi2014one}
V.~Kazemi and J.~Sullivan, ``One millisecond face alignment with an ensemble of
  regression trees,'' in \emph{Proc. IEEE Conf. Comput. Vis. Pattern
  Recognit.}\hskip 1em plus 0.5em minus 0.4em\relax IEEE, 2014, pp. 1867--1874.

\bibitem{guo2018cnn}
Y.~Guo, J.~Zhang, J.~Cai, B.~Jiang, and J.~Zheng, ``{CNN}-based real-time dense
  face reconstruction with inverse-rendered photo-realistic face images,''
  \emph{{IEEE} Trans. Pattern Anal. Mach. Intell.}, 2018.

\bibitem{wei2016dense}
L.~Wei, Q.~Huang, D.~Ceylan, E.~Vouga, and H.~Li, ``Dense human body
  correspondences using convolutional networks,'' in \emph{Proc. IEEE Conf.
  Comput. Vis. Pattern Recognit.}\hskip 1em plus 0.5em minus 0.4em\relax IEEE,
  2016.

\bibitem{chen2014depth}
C.~Chen, H.~X. Pham, V.~Pavlovic, J.~Cai, and G.~Shi, ``Depth recovery with
  face priors,'' \emph{Proc. Asia Conf. Comput. Vis.}, pp. 336--351, 2014.

\bibitem{brunton2014review}
A.~Brunton, A.~Salazar, T.~Bolkart, and S.~Wuhrer, ``Review of statistical
  shape spaces for 3d data with comparative analysis for human faces,''
  \emph{Computer Vision and Image Understanding}, vol. 128, pp. 1--17, 2014.

\bibitem{bouaziz2013online}
S.~Bouaziz, Y.~Wang, and M.~Pauly, ``Online modeling for realtime facial
  animation,'' \emph{ACM Trans. Graph.}, vol.~32, no.~4, p.~40, 2013.

\bibitem{meyer2015robust}
G.~P. Meyer, S.~Gupta, I.~Frosio, D.~Reddy, and J.~Kautz, ``Robust model-based
  3d head pose estimation,'' in \emph{Proc. IEEE Int. Conf. Comput. Vis.}\hskip
  1em plus 0.5em minus 0.4em\relax IEEE, 2015, pp. 3649--3657.

\bibitem{li2015real}
S.~Li, K.~Ngan, R.~Paramesran, and L.~Sheng, ``Real-time head pose tracking
  with online face template reconstruction.'' \emph{{IEEE} Trans. Pattern Anal.
  Mach. Intell.}, 2015.

\bibitem{storer20093d}
M.~Storer, M.~Urschler, and H.~Bischof, ``{3D-MAM}: 3{D} morphable appearance
  model for efficient fine head pose estimation from still images,'' in
  \emph{Proc. IEEE Int. Conf. Comput. Vis. Workshops}.\hskip 1em plus 0.5em
  minus 0.4em\relax IEEE, 2009, pp. 192--199.

\bibitem{6313594}
O.~Aldrian and W.~A.~P. Smith, ``Inverse rendering of faces with a 3d morphable
  model,'' \emph{{IEEE} Trans. Pattern Anal. Mach. Intell.}, vol.~35, no.~5,
  pp. 1080--1093, May 2013.

\bibitem{8010438}
M.~L\"{u}thi, T.~Gerig, C.~Jud, and T.~Vetter, ``Gaussian process morphable
  models,'' \emph{{IEEE} Trans. Pattern Anal. Mach. Intell.}, vol.~40, no.~8,
  pp. 1860--1873, Aug 2018.

\bibitem{tulyakov2014robust}
S.~Tulyakov, R.-L. Vieriu, S.~Semeniuta, and N.~Sebe, ``Robust real-time
  extreme head pose estimation,'' in \emph{Proc. IEEE Int. Conf. Pattern
  Recognit.}\hskip 1em plus 0.5em minus 0.4em\relax IEEE, 2014, pp. 2263--2268.

\bibitem{rusinkiewicz2001efficient}
S.~Rusinkiewicz and M.~Levoy, ``Efficient variants of the icp algorithm,'' in
  \emph{3-{D} Digital Imaging and Modeling, 2001. Proceedings. Third
  International Conference on}.\hskip 1em plus 0.5em minus 0.4em\relax IEEE,
  2001, pp. 145--152.

\bibitem{padeleris2012head}
P.~Padeleris, X.~Zabulis, and A.~A. Argyros, ``Head pose estimation on depth
  data based on particle swarm optimization,'' in \emph{Proc. IEEE Conf.
  Comput. Vis. Pattern Recognit. Workshops}.\hskip 1em plus 0.5em minus
  0.4em\relax IEEE, 2012, pp. 42--49.

\bibitem{wang2016capturing}
R.~Wang, L.~Wei, E.~Vouga, Q.~Huang, D.~Ceylan, G.~Medioni, and H.~Li,
  ``Capturing dynamic textured surfaces of moving targets,'' \emph{Proc. Euro.
  Conf. Comput. Vis.}, 2016.

\bibitem{bolkart2013statistical}
T.~Bolkart and S.~Wuhrer, ``Statistical analysis of 3d faces in motion,'' in
  \emph{Proc. IEEE Conf. 3D Vis.}\hskip 1em plus 0.5em minus 0.4em\relax IEEE,
  June 2013, pp. 103--110.

\bibitem{bishop2006pattern}
C.~M. Bishop and N.~M. Nasrabadi, \emph{Pattern recognition and machine
  learning}.\hskip 1em plus 0.5em minus 0.4em\relax Springer, 2006.

\bibitem{baltruvsaitis20123d}
T.~Baltru{\v{s}}aitis, P.~Robinson, and L.-P. Morency, ``3{D} constrained local
  model for rigid and non-rigid facial tracking,'' in \emph{Proc. IEEE Conf.
  Comput. Vis. Pattern Recognit.}\hskip 1em plus 0.5em minus 0.4em\relax IEEE,
  2012, pp. 2610--2617.

\bibitem{martin2014real}
M.~Martin, F.~Van De~Camp, and R.~Stiefelhagen, ``Real time head model creation
  and head pose estimation on consumer depth cameras,'' in \emph{Proc. IEEE
  Conf. 3D Vis.}, vol.~1.\hskip 1em plus 0.5em minus 0.4em\relax IEEE, 2014,
  pp. 641--648.

\end{thebibliography}


\begin{IEEEbiography}[{\includegraphics[width=1in,height=1.25in,clip]{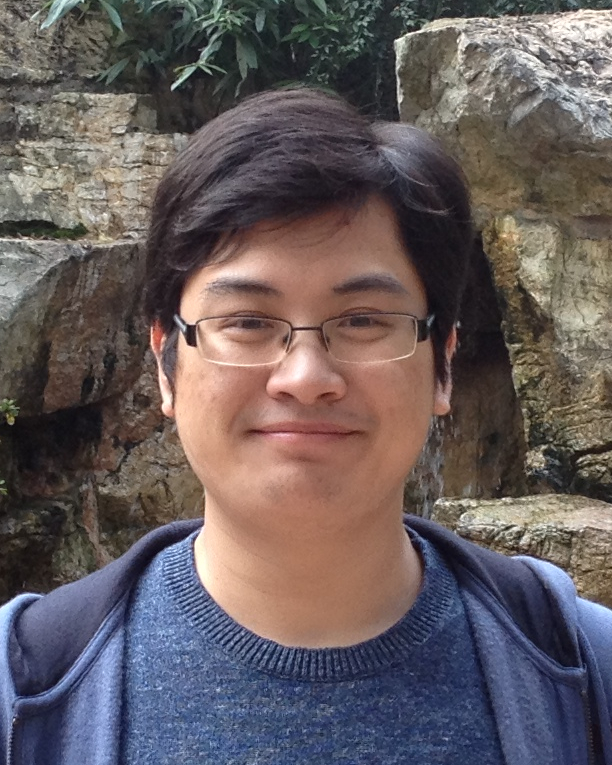}}]{Lu Sheng} (S'13--M'16) is currently an associate professor at the College of Software, Beihang University, China.
Previously, he was a postdoctoral researcher with the Department of Electronic Engineering, the Chinese University of Hong Kong (CUHK).
Before that he received his Ph.D. degree in Electronic Engineering from the Chinese University of Hong Kong in 2016, and B.E. degree in Information Science and Electronic Engineering from Zhejiang University (ZJU) in 2011.
His research interests include deep learning driven low-level and middle-level vision, and 3D vision problems.
\end{IEEEbiography}

\begin{IEEEbiography}[{\includegraphics[width=1in,clip]{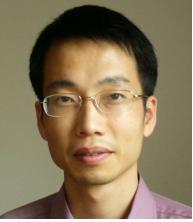}}]{Jianfei Cai}
(S'98--M'02--SM'07) received his PhD degree from the University of Missouri-Columbia.
He is a Professor and has served as the Head of Visual \& Interactive Computing Division and the Head of Computer Communication Division at the School of Computer Science \& Engineering, Nanyang Technological University, Singapore. His major research interests include computer vision, multimedia and deep learning. He has published over 200 technical papers in international journals and conferences. He is currently an Associate Editor for IEEE Trans. on Multimedia, and has served as an Associate Editor for IEEE Trans. on Image Processing and Trans. on Circuit and Systems for Video Technology.
\end{IEEEbiography}


\begin{IEEEbiography}[{\includegraphics[width=1in,clip]{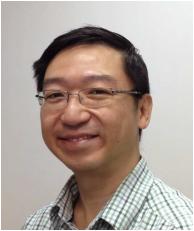}}]{Tat-Jen Cham} is an Associate Professor in the School of Computer Science \& Engineering, Nanyang Technological University, Singapore. After receiving his BA and PhD from the University of Cambridge, he was subsequently a Jesus College Research Fellow, and later a research scientist in the DEC/Compaq Research Lab in Cambridge, MA.

Tat-Jen received overall best paper prizes at PROCAMS2005, ECCV1996 and BMVC1994, and is an inventor on eight patents. He has served as an editorial board member for IJCV, a General Chair for ACCV2014, and Area Chair for past ICCVs and ACCVs.

Tat-Jen’s research interests are broadly in computer vision and machine learning, and he is currently a co-PI in the NRF BeingTogether Centre (BTC) on 3D Telepresence.
\end{IEEEbiography}

\begin{IEEEbiography}[{\includegraphics[width=1in,clip]{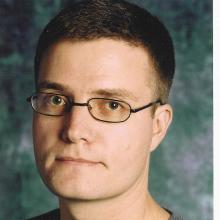}}]{Vladimir Pavlovic}
received the PhD degree in electrical engineering from the University of Illinois
at Urbana-Champaign in 1999. From 1999 to 2001, he was a member of the research staff
at the Cambridge Research Laboratory, Massachusetts. He is an associate professor in the
Computer Science Department at Rutgers University, New Jersey. Before joining Rutgers in
2002, he held a research professor position in the Bioinformatics Program at Boston University. His research interests include probabilistic system modeling, time-series analysis, computer vision, and bioinformatics. He is a senior member of the IEEE.
\end{IEEEbiography}

\begin{IEEEbiography}[{\includegraphics[width=1in,height=1.25in,clip]{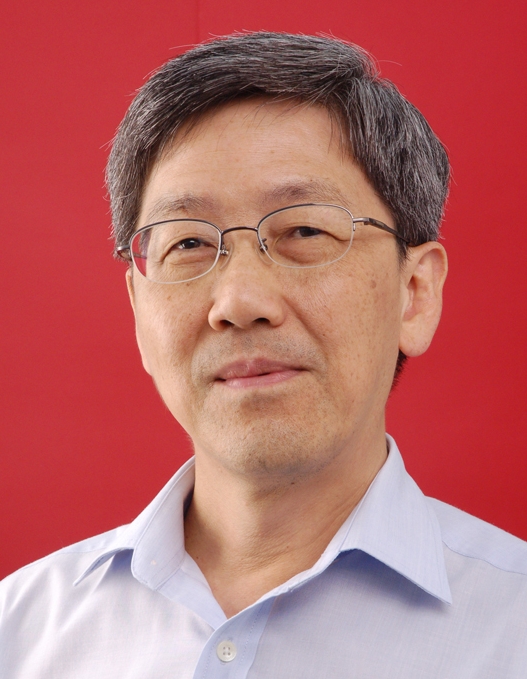}}]{King Ngi Ngan}
(F'00) King N. Ngan (F’00) received the Ph.D. degree in Electrical Engineering from the Loughborough University in U.K.  He is currently a Chair Professor at the University of Electronic Science and Technology, Chengdu, China, under the National Thousand Talents Plan. He was previously a Chair Professor at the Chinese University of Hong Kong, the Nanyang Technological University, Singapore, and the University of Western Australia, Australia. He holds honorary and visiting professorships of numerous universities in China, Australia and South East Asia.

Prof. Ngan served as associate editor of IEEE Transactions on Circuits and Systems for Video Technology, Journal on Visual Communications and Image Representation, EURASIP Journal of Signal Processing: Image Communication, and Journal of Applied Signal Processing.  He chaired and co-chaired a number of prestigious international conferences on image and video processing including the 2010 IEEE International Conference on Image Processing, and served on the advisory and technical committees of numerous professional organizations.  He has published extensively including 3 authored books, 7 edited volumes, over 400 refereed technical papers, and edited 9 special issues in journals. In addition, he holds 15 patents in the areas of image/video coding and communications. 

Prof. Ngan is a Fellow of IEEE (U.S.A.), IET (U.K.), and IEAust (Australia), and an IEEE Distinguished Lecturer in 2006-2007.
\end{IEEEbiography}


%\vfill


\end{document}